\documentclass[journal]{new-aiaa}
\usepackage[utf8]{inputenc}
\usepackage{textcomp}

\usepackage{graphicx}
\usepackage{amsmath}
\usepackage[version=4]{mhchem}
\usepackage{siunitx}
\usepackage{longtable,tabularx}

\usepackage{array}
\usepackage{longtable}%tabulary
\usepackage{subcaption} % for subfigures
\usepackage{placeins} % for FloatBarrier
\usepackage{tikz} % for the NF illustration
\usetikzlibrary{shapes.geometric, arrows}

\newcommand{\argminE}{\mathop{\mathrm{argmin}}}    

\title{Virtual Target Trajectory Prediction for Stochastic Targets}

\author{Marc Schneider\footnote{Research Associate, Institute of Flight Mechanics and Controls, corresponding author. Email: marc.schneider@ifr.uni-stuttgart.de}, Renato Loureiro\footnote{Research Associate, Institute of Flight Mechanics and Controls.}, Torbjørn Cunis\footnote{Lecturer, Institute of Flight Mechanics and Controls, AIAA Member.}, Walter Fichter\footnote{Professor, Institute of Flight Mechanics and Controls, AIAA Associate Fellow.}}
\affil{University of Stuttgart, 70569 Stuttgart, Germany}

\begin{document}

\maketitle
\footnotetext{Substantial parts presented as Paper CEAS-GNC-2024-103 at the CEAS EuroGNC 2024, Bristol, UK, 11-13 June 2024.}

\begin{abstract}
Trajectory prediction of aerial vehicles is a key requirement in applications ranging from missile guidance to UAV collision avoidance. 
While most prediction methods assume deterministic target motion, real-world targets often exhibit stochastic behaviors such as evasive maneuvers or random gliding patterns. 
This paper introduces a probabilistic framework based on Conditional Normalizing Flows (CNFs) to model and predict such stochastic dynamics directly from trajectory data. 
The learned model generates probability distributions of future target positions conditioned on initial states and dynamic parameters, enabling efficient sampling and exact density evaluation. 
To provide deterministic surrogates compatible with existing guidance and planning algorithms, sampled trajectories are clustered using a time series k-means approach, yielding a set of representative “virtual target” trajectories. 
The method is target-agnostic, computationally efficient, and requires only trajectory data for training, making it suitable as a drop-in replacement for deterministic predictors. 
Simulated scenarios with maneuvering and ballistic targets demonstrate that the proposed approach bridges the gap between deterministic assumptions and stochastic reality, advancing guidance and control algorithms for autonomous vehicles.
\end{abstract}

\section{Introduction}
\lettrine{P}{rediction} models play a critical role in the guidance and control of guided missiles. 
In order to effectively guide a missile to its target, it is necessary to predict the future trajectory of both the missile and the target. 
These predictions enable the computation of control commands within a guidance law. 
While the trajectory of the missile can often be predicted with high accuracy due to the deterministic nature of its dynamics, predicting the trajectory of the target is much more challenging because of its inherently stochastic behavior.

Traditional prediction models for target motion assume deterministic behavior, either implicitly or explicitly. 
However, relying on a deterministic model for the target can degrade the performance of the guidance law because the control commands are based on a fixed prediction of the trajectory that may not align with the actual, more varied maneuvers that the target is executing. 
In reality, the motion of the target is not confined to a single deterministic model; it is capable of executing a wide range of possible maneuvers.

This shift towards a probabilistic framework presents new challenges, particularly in the prediction of the future trajectory of the target, which is now described by a dynamic probability distribution rather than a single deterministic path. 
The future states of the target cannot be described as distinct trajectories, but rather as a time-varying probability distribution over the state space. 
Such problems are called multimodal trajectory prediction problems and have mostly been studied in the context of autonomous driving and pedestrians, but not in the context of guided missiles.

One challenge for this kind of problem is that the probability distribution over the state space usually cannot be derived analytically, but rather has to be approximated in some way. 
Moreover, most algorithms (guidance laws, collision avoidance, etc.) expect deterministic trajectories, which creates a gap between the probabilistic nature of target motion and the requirements of downstream applications.

Many traditional approaches have attempted to address target motion prediction. 
For example, the Proportional Navigation (PN) guidance law~\cite{adler1956missile} assumes that the target maintains a constant velocity, while the Zero-Effort-Miss (ZEM) guidance law~\cite{nesline1981new,guo2013applications} allows for arbitrary target dynamics within a deterministic framework.

Recognizing the limitations of purely deterministic approaches,~\cite{Dwivedi2016} introduced an approach using an Extended Kalman Filter (EKF) to estimate not only the position and velocity of the target, but also the parameters governing its dynamics. 
This concept was further advanced in~\cite{schneider2022multi}, where a multi-hypothesis guidance approach was proposed using an Interacting Multiple Model (IMM) filter to estimate both the state of the target and the probability of various dynamics hypotheses.

While this multi-hypothesis guidance technique is powerful, it has two significant drawbacks.
First, the number of hypotheses that can be considered is limited by computational resources.
Second, despite incorporating multiple hypotheses, the approach still assumes that the target follows one of several deterministic trajectories, which may not fully capture the stochastic nature of target motion.

In~\cite{shaviv2006estimation}, a Monte Carlo method is used to approximate the distribution of future positions of a target, but due to computational demands, only a few samples can be generated.
Ref.~\cite{huang2023multimodal} gives a good overview of different approaches to multimodal trajectory prediction, ranging from adding noise to deterministic predictions, anchor methods~\cite{chai2019multipath}, clustering and Gaussian Mixture Model approaches~\cite{barratt2018learning}, grid-based methods~\cite{guo2022end}, to generative Machine Learning (ML) techniques.

ML approaches include Generative Adversarial Networks (GANs)~\cite{goodfellow2020generative}, where a classifier is trained to distinguish between real data and data generated by a generator network.
In a minimax game, the generator network is trained to generate trajectories that are indistinguishable from real trajectories.
Variational Autoencoders (VAEs)~\cite{kingma2013auto} allow for the generation of new samples by encoding input data into a latent space and then decoding it back to the original space.
Diffusion models~\cite{ho2020denoising} are a recent class of generative models that have shown great success in generating high-quality samples.
However, diffusion models typically require hundreds of function evaluations for inference, making them impractical for real-time applications like missile guidance.
In~\cite{meszaros2024trajflow}, a diffusion model is compared to other methods for trajectory prediction, performing significantly worse than them.

Normalizing Flows (NFs)~\cite{rezende2015variational,papamakarios2021normalizing} are a class of generative ML models that can be used to approximate
probability distributions by transforming a simple base distribution into the desired, complex distribution and vice versa
using a series of invertible transformations. Compared to other ML approaches like GANs or VAEs, NFs have the
advantage that they can perform both sampling (of possible future states) and inference (i.e., calculating the probability
of a given state). Moreover, they allow for efficient and exact computation of the probability of a given state, giving rise
to heatmap-like visualizations of the probability density function (PDF) over the state space.

In~\cite{scholler2021flomo} Conditional NFs (CNFs) are used to predict the future trajectories of pedestrians in a multimodal way.
It employs recurrent neural networks to encode the past trajectory of the pedestrians which is used to calculate the conditional probability distribution over the future trajectory conditioned on the past trajectory. Each predicted trajectory consists of a sequence of positions with equal time intervals between them. 

In \cite{ma2021likelihood} a method is proposed to improve the quality and diversity of trajectories generated by NFs.
By adding a diversity objective function, more diverse predictions for the future trajectory of vehicles with discrete time steps can be predicted conditioned on measurements and additional physical attributes.
To accommodate the need for fast inference,~\cite{maeda2023fast} proposes a method to speed up the inference of NFs by reusing the results of previous computations to predict future trajectories of humans.

Only the method in our previous work~\cite{CEAS-GNC-2024-103} predicts probability distributions of future positions for arbitrary times using CNFs.
With the CNFs approach, the distribution of the future position of stochastically moving targets can be predicted.
The advantage of this approach is that the predictions in the form of a PDF can be interpreted and visualized as a heatmap, which can be used to gain insights into the predicted behavior of the target.
However, almost all algorithms (guidance laws, collision avoidance, ...) expect deterministic trajectories, which is a simplification that might not hold in reality.
To bridge this gap, this paper employs a clustering algorithm to cluster the samples generated by the CNFs into a set of representative trajectories, called virtual targets.
Since the samples can be generated for equal time steps with equal sample sizes, most of the problems that arise when clustering trajectory data can be avoided.
Thus, a time series k-means clustering algorithm~\cite{macqueen1967some} is used to generate trajectories of the virtual targets.
These virtual targets can then be used as a drop-in replacement for deterministic trajectory predictions in guidance laws, path planning, or other applications that require deterministic trajectory predictions.
The contributions of this paper are two-fold:

\begin{enumerate}
\item A prediction framework leveraging CNFs to predict the distribution of future positions of stochastically moving targets, allowing for interpretable, target-agnostic, and fast predictions.

\item A time series k-means clustering algorithm to cluster samples generated by CNFs into representative trajectories, called virtual target trajectories, which can be used as a drop-in replacement for deterministic trajectory predictions for various applications.
\end{enumerate}
% structure of the paper
The remainder of this paper is structured as follows:
In Sec.~\ref{sec:problem_statement}, the problem of trajectory prediction for stochastic targets is described mathematically.
Sec.~\ref{sec:NFs} explains the theory of NFs and describes their application to the problem at hand.
The subsequent clustering process to calculate the trajectories of the virtual targets is described in Sec.~\ref{sec:generation}.
These process is illustrated in Fig.~\ref{fig:structure}.
The generation of the training data is described in Sec.~\ref{sec:data_generation}.
Sec.~\ref{sec:results} presents the results of the approach and Sec.~\ref{sec:conclusion} concludes the paper.

\tikzstyle{startstop} = [rectangle, rounded corners, minimum width=3cm, minimum height=1cm,text centered, draw=black, fill=red!30]
\tikzstyle{process} = [rectangle, minimum width=3cm, minimum height=1cm, text centered, draw=black, fill=orange!30]
\tikzstyle{decision} = [diamond, minimum width=3cm, minimum height=1cm, text centered, draw=black, fill=green!30]
\tikzstyle{arrow} = [thick,->,>=stealth]

\begin{figure}[h!]
    \centering
    \begin{tikzpicture}[node distance=2cm, scale=1.0]
        % Nodes
        \node (start) [startstop] {Input: Positions and Velocities of Targets};
        \node (cnf) [process, below of=start, node distance=2cm] {
            \begin{tabular}{c}
                \begin{minipage}{0.38\textwidth}
                    Sample Future Target Positions From Conditional Normalizing Flows (Sec.~\ref{sec:NFs}) \\
                \end{minipage}%
            \end{tabular}
        };
        \node (translate_rotate) [process, below of=cnf, node distance=2cm] {
            \begin{tabular}{c}
                \begin{minipage}{0.38\textwidth}
                    Translate and Rotate Samples (Sec.~\ref{sec:NFs}) \\
                \end{minipage}%
            \end{tabular}
        };
        \node (clustering) [process, below of=translate_rotate, node distance=2cm] {
            \begin{tabular}{c}
                \begin{minipage}{0.38\textwidth}
                    Time Series K-Means Clustering (Sec.~\ref{sec:generation}) \\
                \end{minipage}%
            \end{tabular}
        };
        \node (stop) [startstop, below of=clustering, node distance=2cm] {Output: Cluster Centers (Virtual Target Trajectories)};

        % Arrows
        \draw [arrow] (start) -- (cnf);
        \draw [arrow] (cnf) -- (translate_rotate);
        \draw [arrow] (translate_rotate) -- (clustering);
        \draw [arrow] (clustering) -- (stop);
    \end{tikzpicture}
    \caption{Illustration of the structure of the prediction framework.}
    \label{fig:structure}
\end{figure}
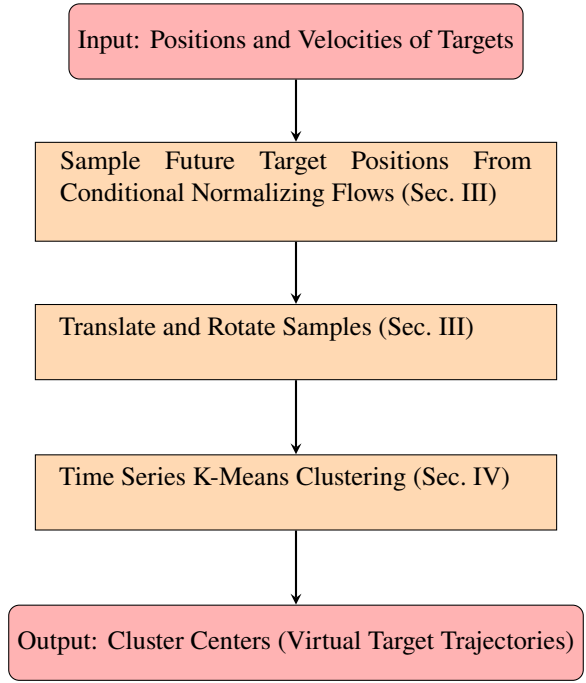

\section{Problem Statement}
\label{sec:problem_statement}
The main problem to be solved in this paper is the prediction of the future trajectory of the stochastically maneuvering target in a computationally efficient manner and its meaningful representation for downstream tasks that require deterministic trajectory predictions.
It is assumed that either the stochastic model of the target dynamics is known or a dataset of target trajectories is available.

The future position $\mathbf{x}(t)$ at time $t$ of a moving target performing random maneuvers is to be predicted given the initial position $\mathbf{x}_0$ at $t=0$ and additional parameters $\mathbf{\psi}$ regarding the target dynamics, such as the ballistic coefficient or the maximum turn rate.
More precisely, the PDF $p(\mathbf{x} \mid t, \mathbf{x}_0, \mathbf{\psi})$ is to be estimated, which describes the relative probability of the target being at a given position $\mathbf{x}$ at time $t$ conditioned on the initial position $\mathbf{x}_0$ and the parameters $\mathbf{\psi}$.

Additionally, an algorithm $C_{n_v}$ to generate a set of $n_\mathrm{v}$ representative trajectories $\phi_i$ from set of PDFs $p_j(\mathbf{x} \mid t, \mathbf{x}_0, \mathbf{\psi})$ for $n_\mathrm{r}$ targets is to be developed, such that each trajectory $\phi_i = \{\mathbf{x}_{i}(t_k)\}_{k=1}^{n_t}$ consisting of the sequence of $n_t$ positions $\mathbf{x}_{i}$ at times $t_k$ are generated for each virtual target $i$.
Equation~\eqref{eq:problem} describes the mapping of the input data to the output data of the CNFs model.

\begin{equation}
    \label{eq:problem}
    C_{n_v}: \{p_j(\mathbf{x} \mid t, \mathbf{x}_0, \mathbf{\psi})\}_{j=1}^{n_r} \mapsto \{\phi_i\}_{i=1}^{n_v}
\end{equation}

\section{Learning Stochastic Target Dynamics}
\label{sec:NFs}
The first step to solve the problem described in Sec.~\ref{sec:problem_statement} is to learn the stochastic behavior of the target.
In order to learn the probability distribution of the target's future position, NFs are used.
NFs are a class of generative ML models that can be used to approximate probability distributions by transforming a simple base distribution into the desired, complex distribution and vice versa.
Here, the complex distribution is the distribution of the position of the target at a given time and the base distribution is a Gaussian distribution.
In the following, we first provide background on the established theory of NFs and CNFs, before describing their novel application to the problem at hand.

\subsection{Normalizing Flows}
NFs are an established class of generative ML models introduced by~\cite{rezende2015variational} that consist of a series of invertible transformations, which transform a sample from the base distribution into a sample from the complex distribution and vice versa.
They consist of a series of invertible transformations, which transform a sample from the base distribution into a sample from the complex distribution and vice versa.
Since the transformations are invertible, the model can be used for both sampling and inference. 
Sampling means that the NFs model can be used to generate samples from the desired distribution by sampling from the base distribution and transforming the samples with the learned invertible transformations.
Inference, or density estimation, means that the model can be used to calculate the probability density of a given sample from the complex distribution by transforming it to the base distribution and calculating the probability of the sample in the known base distribution and correcting for the volume change due to the transformation by multiplying with the determinant of the Jacobian of the transformation.

\begin{figure}
    \centering
    \resizebox{0.8\textwidth}{!}{
        \includegraphics{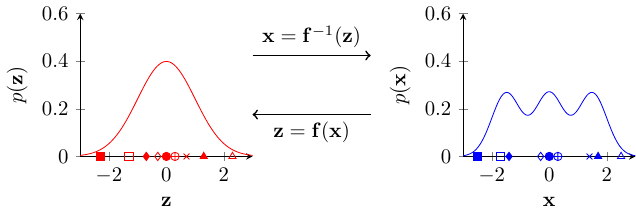}
    }
    \caption{Illustration of the transformation of samples from a normal distribution to a complex distribution (and vice versa) using Normalizing Flows. Figure adapted from Schneider, M., Loureiro, R., Cunis, T., and Fichter, W., “Trajectory Prediction for Missile Targets: A Probabilistic Approach Using Machine Learning,” in CEAS EuroGNC Conference, 2024. Licensed under a Creative Commons Attribution 4.0 International License (CC-BY 4.0).}
    \label{fig:NF_illustration}
\end{figure}

Figure~\ref{fig:NF_illustration} (adapted from~\cite{CEAS-GNC-2024-103}) illustrates the transformation of samples from a normal distribution to a complex distribution (and vice versa) using NFs.
The samples are depicted by the shapes (squares, circles, \ldots) and their relative probability is indicated by the probability density function above.
In general, NFs describe a bijection $\mathbf{f}$, which transforms samples $\mathbf{x} \in \mathbb{R}^d$ from the complex distribution $p(\mathbf{x})$ into samples $\mathbf{z} \in \mathbb{R}^d$ from the base distribution $p(\mathbf{z})$.
In this application, $p(\mathbf{z})$ is a Gaussian distribution with zero mean and unit variance and $p(\mathbf{x})$ denotes the distribution of the position $\mathbf{x}$ of the target.

Equation~\eqref{eq:NFs} describes the transformation $\mathbf{f}$ and its inverse $\mathbf{f}^{-1}$ as a composition of $n_\mathrm{l}$ invertible transformations $\mathbf{f}_{\Theta_i}$, where $\Theta_i$ are the $n_{\Theta_i}$ parameters of the $i$-th transformation.
The transformations $\mathbf{f}_{\Theta_i}$ can be any invertible function, provided they are easy to evaluate, invert, and differentiate.
Moreover, the determinant of the Jacobian of the transformation $\mathbf{f}_{\Theta_i}$ must be easy to compute since it is needed for the calculation of the PDF of the complex distribution.
Examples of such transformations include affine transformations (e.g., $\mathbf{f}_{\Theta_i}(\mathbf{x}) = \mathbf{A} \mathbf{x} + \mathbf{b}$), as utilized in RealNVP~\cite{dinh2016density}.
In this paper, $\mathbf{f}_{\Theta_i}$ are rational quadratic spline transformations, which are elaborated upon in Sec.~\ref{sec:application}.
The parameters $\Theta_i$ are the outputs of neural networks (NNs) $\alpha_i$, which are trained to learn the parameters of the transformations.
\begin{align}
    \mathrm{NFs}: \quad & 
    \begin{cases}
        \mathbf{z} = \mathbf{f}(\mathbf{x}) = (\mathbf{f}_{\Theta_{n_\mathrm{l}}} \circ \dots \circ \mathbf{f}_{\Theta_2} \circ \mathbf{f}_{\Theta_1})(\mathbf{x}) \\
        \mathbf{x} = \mathbf{f}^{-1}(\mathbf{z}) = (\mathbf{f}_{\Theta_1}^{-1} \circ \dots \circ \mathbf{f}_{\Theta_{n_\mathrm{l}}}^{-1})(\mathbf{z}) \\
        \text{with } \Theta_i = \alpha_i(\mathbf{f}_{\Theta_{i-1}}(\mathbf{x}))
        \label{eq:NFs}
    \end{cases}
\end{align}

\begin{figure}
    \centering
    \includegraphics{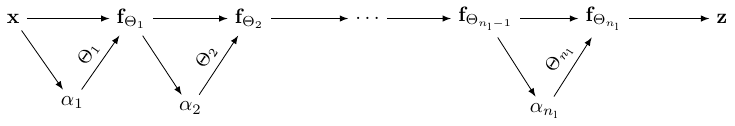}
    \caption{Illustration of the Normalizing Flows concept with neural networks $\alpha_i$ and transformations $f_{\Theta_i}$ with parameters $\Theta_i$. Figure adapted from Schneider, M., Loureiro, R., Cunis, T., and Fichter, W., “Trajectory Prediction for Missile Targets: A Probabilistic Approach Using Machine Learning,” in CEAS EuroGNC Conference, 2024. Licensed under a Creative Commons Attribution 4.0 International License (CC-BY 4.0).}
    \label{fig:NF_concept}
\end{figure}

Equation~\eqref{eq:alpha_i} depicts the input and output dimensions of the NNs $\alpha_i$ with $d$ being the dimension of the input data and $n_{\Theta_i}$ being the size of the parameters of the transformation $\mathbf{f}_{\Theta_i}$.
The input for $\alpha_i$ is the output from the previous transformation $\mathbf{f}_{\Theta_{i-1}}$ as illustrated in Fig.~\ref{fig:NF_concept} (adapted from~\cite{CEAS-GNC-2024-103}).

\begin{equation}
    \label{eq:alpha_i}
    \alpha_i: \mathbb{R}^{d} \rightarrow \mathbb{R}^{n_{\Theta_i}}
\end{equation}

Using samples from the complex distribution, the weights of the NNs are optimized during the training process in a Maximum Likelihood Estimation approach, such that the observed training data points are most probable under the learned distribution.
This is done by minimizing the negative log-likelihood of the training data points, which is equivalent to maximizing the likelihood of creating the training data points by sampling from the base distribution and transforming them with the NFs.

Thus, the optimal set of weights $\theta^*$ for the NNs $\alpha_l$ with $\theta = \{\theta_1, \ldots, \theta_{n_\mathrm{l}}\}$ is found by minimizing the negative log-likelihood of the $n_d$ training data points $\mathbf{x}_i$ when applying the NFs $p_\theta(\mathbf{x})$ with NN weights and biases $\theta$ as described in Eq.~\eqref{eq:training}:

\begin{equation}
    \label{eq:training}
    \theta^* = \arg\min_{\theta} \left(-\sum_{i=1}^{n_\mathrm{d}} \log p_\theta(\mathbf{x}_i) \right)
\end{equation}

For simplicity, the index $\theta$ indicating the set of NN weights and biases in the NFs model is dropped in the following.
During the training process, singularities of the true PDF $p(\mathbf{x})$, i.e., points where the PDF has a value of infinity, can lead to problems.
The reason for this is that it would violate the invertibility property since the inverse transformation $\mathbf{f}^{-1}$ would not be defined at these points.
A common solution to this problem, which is also applied in this paper, is the so-called noise injection, where a small amount of noise is added to the training data to dilute the singularities of the true PDF.

\subsection{Conditional Normalizing Flows}
The NFs model can be used to transform samples from the complex distribution to the base distribution (density estimation) and to sample from the complex distribution.
However, a NFs model can only transform a simple base distribution into one complex distribution, i.e., for one time $t$ and one set of parameters $\mathbf{\psi}$.
Since the model should not describe only one distribution, but rather a distribution for any given time $t$ and additional parameters $\mathbf{\psi}$ regarding the target dynamics, the architecture must be extended.
This can be achieved by using Conditional Normalizing Flows (CNFs)~\cite{papamakarios2021normalizing}, which are a special kind of NFs that can be conditioned on some input to calculate a conditional probability $p(\mathbf{x} \mid t, \mathbf{\psi})$ in contrast to the unconditional probability $p(\mathbf{x})$ provided by normal NFs.
In this case, $t$ and $\mathbf{\psi}$ serve as additional input for the model, the so-called conditioning variables.
More precisely, they are used as an additional input for the NNs besides the output of the previous transformation $\mathbf{f}_{\Theta_{i-1}}$.
To differentiate these NNs with more inputs from the NNs $\alpha_i$ used in the NFs model, they are denoted as $\beta_i$.

\begin{figure}
    \centering
    \includegraphics{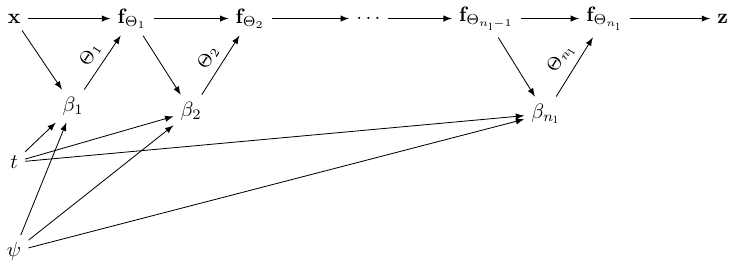}
    \caption{Illustration of the Conditional Normalizing Flows concept with neural networks $\beta_i$ and transformations $f_{\Theta_i}$ with parameters $\Theta_i$, conditioned on the time $t$ and dynamics parameters $\psi$.}
    \label{fig:CNF_concept}
\end{figure}

As displayed in Fig.~\ref{fig:CNF_concept}, CNFs describe a transformation $\mathbf{f}$, which transforms samples $\mathbf{x}$ from the complex distribution $p(\mathbf{x} \mid t, \mathbf{\psi})$ into samples $\mathbf{z}$ from the base distribution $p(\mathbf{z})$.
Equation~\eqref{eq:CNFs} describes the transformation $\mathbf{f}$ and its inverse $\mathbf{f}^{-1}$ as a composition of $n_\mathrm{l}$ invertible transformations $\mathbf{f}_{\Theta_i}$, where $\Theta_i$ are the $n_{\Theta_i}$ parameters of the $i$-th transformation.
The parameters $\Theta_i$ are the outputs of NNs $\beta_i$, which are trained to learn the parameters of the transformations conditioned on $t$ and $\mathbf{\psi}$.
\begin{align}
    \mathrm{CNFs}: \quad &
    \begin{cases}
        \mathbf{z} = \mathbf{f}(\mathbf{x}, t, \mathbf{\psi}) = (\mathbf{f}_{\Theta_{n_\mathrm{l}}} \circ \dots \circ \mathbf{f}_{\Theta_2} \circ \mathbf{f}_{\Theta_1})(\mathbf{x}, t, \psi) \\
        \mathbf{x} = \mathbf{f}^{-1}(\mathbf{z}, t, \mathbf{\psi}) = (\mathbf{f}_{\Theta_1}^{-1} \circ \dots \circ \mathbf{f}_{\Theta_{n_\mathrm{l}}}^{-1})(\mathbf{z}) \\
        \text{with } \Theta_i = \beta_i(\mathbf{f}_{\Theta_{i-1}}(\mathbf{x}), t, \mathbf{\psi})
    \end{cases}
    \label{eq:CNFs}
\end{align}
Equation~\eqref{eq:beta_i} depicts the input and output dimensions of the NNs $\beta_i$.
The input for $\beta_i$ is the output from the previous transformation $\mathbf{f}_{\Theta_{i-1}}$, the time $t$, and the parameters $\mathbf{\psi}$ as illustrated in Fig.~\ref{fig:CNF_concept}.
\begin{equation}
    \label{eq:beta_i}
    \beta_i: \mathbb{R}^{d + 1 + n_{\psi}} \rightarrow \mathbb{R}^{n_{\Theta_i}}
\end{equation}

\subsection{Application of Conditional Normalizing Flows}
\label{sec:application}
In our approach, we utilize CNFs to model the distribution of the target's position under varying conditions of time and target dynamics. Specifically, we implement CNFs with a technique known as Masked Autoregressive Flow (MAF) combined with rational quadratic splines.

MAF~\cite{papamakarios2017masked} is a method that leverages autoregressive models to generate samples from a complex distribution. It does so by modeling the distribution as a sequence of conditional distributions where each dimension depends on the previous ones, giving it a high degree of flexibility.
This autoregressive property makes it suitable for our task of predicting the target's position under different conditions.

Rational quadratic splines~\cite{durkan2019neural} are used as transformation functions $\mathbf{f}_{\Theta_i}$ to further enhance the flexibility and expressive power of the model compared to affine transformations as used in RealNVP~\cite{dinh2016density}.
They allow to capture complex patterns in the data by adjusting the shape of the spline as needed, making the model adaptable to a wide range of target distributions.

While a detailed technical description of MAF and rational quadratic splines is beyond the scope of this paper, we refer the interested reader to the original papers for a more in-depth understanding of these techniques.
The combination of MAF and rational quadratic splines in CNFs allows for a flexible and expressive model that can accurately capture the distribution of the target's position.

Leveraging CNFs allows to transform samples from the complex distribution to the base distribution for a given time $t$ and additional parameters $\mathbf{\psi}$ regarding the target dynamics which serve as the conditioning variables.
Thus, the PDF $p(\mathbf{x} \mid t, \mathbf{x}_0, \mathbf{\psi})$ can be evaluated by transforming a position $\mathbf{x}$ from the complex distribution to the base distribution and calculating the probability of the sample in the base distribution.
Moreover, the model can be used to sample from the complex distribution for a given time $t$ and parameters $\mathbf{\psi}$ by sampling from the base distribution and transforming the samples with the learned invertible transformations.
The calculation time for the evaluation of the PDF or sampling is constant, since the model learns the distribution of the target's future position and not the dynamics of the target itself.
Not only is the sampling using the CNFs much faster than Monte Carlo simulation (especially for evaluations at the final time), the use of CNFs also allows for an exact computation of the PDF, which is not possible with the Monte Carlo trajectory generation method. 
To facilitate the training process, the training data is normalized to a range of $[-1, 1]$, including the positions $\mathbf{x}$, the time $t$, and the parameters $\mathbf{\psi}$.

\subsection{Translation and Rotation of the Predicted Target Positions}
\label{sec:translation_rotation}
Since the true distribution is independent of the absolute position and the rotation of the target for the scenarios used in this paper, the position of the target can be predicted for any initial position and orientation of the target by transforming the position of the target to the origin and rotating it such that the target is flying northbound.
Thus, the prediction task can be simplified from predicting $p(\mathbf{x} \mid t, \mathbf{x}_0, \mathbf{\psi})$ to predicting $p(\mathbf{x} \mid t, \mathbf{\psi})$.
This simplified PDF is then translated to the current normalized position $\bar{\mathbf{p}}$ and rotated according to the orientation of the target.
The position $\bar{\mathbf{p}}$ is calculated by applying the same normalization procedure to the current position $\mathbf{p}$ as was applied to the training data.

\begin{figure}
    \centering
    \includegraphics{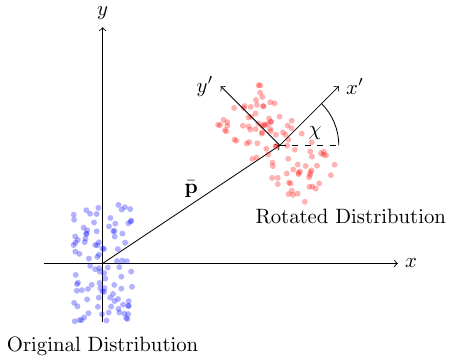}
    \caption{Illustration of the translation (by $\bar{\mathbf{p}}$) and the rotation (by $\chi$) of the predicted target positions.}
    \label{fig:translation_rotation}
\end{figure}

In a two-dimensional scenario, the rotation matrix $\mathbf{R}$ is calculated from the orientation $\chi$ of the target as described in Eq.~\eqref{eq:rotation_matrix}.
For a three-dimensional scenario, the extension is straightforward.

\begin{equation}
    \mathbf{R} = \begin{bmatrix}
        \cos(\chi) & -\sin(\chi) \\
        \sin(\chi) & \cos(\chi)
    \end{bmatrix}
    \label{eq:rotation_matrix}
\end{equation}
The samples $\mathbf{x}$ predicted by the CNFs are then moved to the normalized position $\bar{\mathbf{p}}$ of the target by performing the rotation and translation as described in Eq.~\eqref{eq:translation_rotation} and visualized in Fig.~\ref{fig:translation_rotation}.
Note that these samples $\mathbf{x}_{\mathrm{real}}$ are still in the normalized coordinate system.
\begin{equation}
    \mathbf{x}_{\mathrm{real}} = \mathbf{R} \mathbf{x} + \bar{\mathbf{p}}
    \label{eq:translation_rotation}
\end{equation}
Nevertheless, if the true target distribution depends on the initial position and orientation, i.e., if the motion of the target is not translation and rotation invariant, the initial position and orientation of the target can be used as additional conditioning variables to allow for the prediction of the position of the target for any initial position and orientation.

\subsection{Outlier Removal}
\label{sec:outlier_removal}
Due to the normalization of the training data, the domain that the CNFs are trained on is limited to $[-1, 1]$.
Thus, the performance of the CNFs outside this domain can be arbitrarily poor.
Furthermore, these samples might even be physically impossible, e.g., if the velocity of the target is assumed to be constant.
To prevent the usage of such outliers, samples outside this range plus a margin of three times the standard deviation of the noise injection are removed from the set of generated samples.

\section{Generation of Virtual Target Trajectories}
\label{sec:generation}
While the CNFs can be used to predict the distribution of the future positions of the target, the samples generated by the CNFs are not directly usable for downstream tasks, since they are not deterministic trajectories.
Besides the development of guidance laws that can handle stochastic trajectories, a more straightforward approach is to generate a set of representative trajectories from the samples, which can be used as a drop-in replacement for deterministic trajectory predictions.
To this end, a time series k-means clustering algorithm is used to cluster the samples into a set of representative trajectories.
This approach, i.e., the combination of the predictions for multiple targets and the clustering of the samples, as well as the renormalization of the cluster means, is described in the following sections.
Figure~\ref{fig:structure} illustrates the structure of the prediction framework, which consists of three main steps: sampling from the CNFs, translating and rotating the samples, and clustering the samples into a set of representative trajectories.

\subsection{Combination of Targets}
\label{sec:combination}
For each of the $n_\mathrm{r}$ real targets $i = 1, \ldots, n_\mathrm{r}$, the respective learned CNFs model is used to generate $j=1, \ldots, n_\mathrm{s}$ samples $\mathbf{x}_{i,j,k}$ of the future positions of the target $i$ for each time step $k = 1, \ldots, n_t$.
First, $n_\mathrm{r}  n_\mathrm{s}$ samples $\mathbf{z}_{i,j}$ are drawn from the base distribution, which is a Gaussian distribution with zero mean and unit variance of dimension $d$, i.e.
\begin{equation}
    \mathbf{z}_{i,j} \sim \mathcal{N}(\mathbf{0}, \mathbf{I}_d) \quad \text{for } i = 1, \ldots, n_\mathrm{r},~ j = 1, \ldots, n_\mathrm{s}
\end{equation}
Then, the samples are transformed by the CNFs $\mathbf{f}$ to obtain samples from the complex distribution:
\begin{equation}
    \mathbf{x}_{i,j,k} = \mathbf{f}^{-1}(\mathbf{z}_{i,j}, t_k, \mathbf{\psi}_i) \quad \text{for } i = 1, \ldots, n_\mathrm{r}, ~j = 1, \ldots, n_\mathrm{s}, ~k = 1, \ldots, n_t
\end{equation}
$\mathbf{x}_{i,j,k}$ consists of $d$ dimensions, where $d$ is the dimension of the position of the target which can be 2 or 3 depending on the scenario.
If the targets are of the same type, i.e., their dynamics and their parameters $\mathbf{\psi}$ are identical, the same generated samples can be used for all the targets to save computation time.
Otherwise, a separate CNFs model has to be trained for each type of target and then evaluated for each target.
Since most target distributions are not dependent on the current position and azimuth, the samples from the CNFs which are generated in a normalized coordinate system with axis limits of $[-1, 1]$ can be moved to the normalized current position $\bar{\mathbf{p}}_i$ with orientation $\chi_i$ of the real target by performing a rotation and a translation as described in Sec.~\ref{sec:translation_rotation}.

\subsection{Clustering}
\label{sec:clustering}
\begin{figure}
    \centering
    \includegraphics{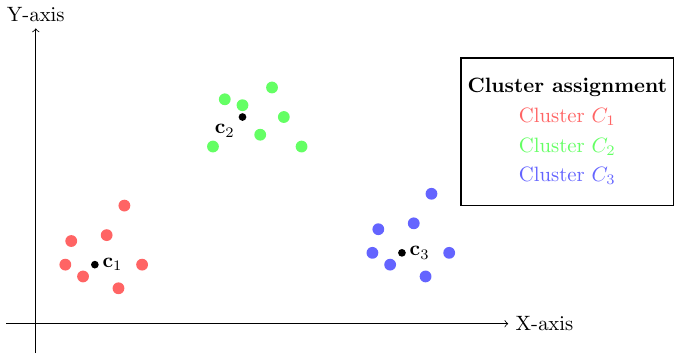}
    \caption{Illustration of the result of k-means clustering for two dimensions.}
    \label{fig:kmeans_clustering}
\end{figure}

After generating and then rotating and translating the samples or each real target $i$, the trajectories $\mathbf{y}_{i,j}$ of the samples are clustered to obtain a set of representative trajectories.
To this end, a time series k-means clustering algorithm is used, which assigns each sample trajectory to one of $n_\mathrm{v}$ clusters, where $n_\mathrm{v}$ is the number of virtual targets that can be chosen by the user.
The resulting trajectories of the cluster means are then used as the virtual targets.
The input for the clustering process is a set $Y$ comprised of $n_\mathrm{r}  n_\mathrm{s}$
flattened trajectories $\mathbf{y}_{i,j}$ of size ($n_t  d \times 1$), which are generated as follows:

\begin{equation}
    \mathbf{y}_{i,j} = \left[ \mathbf{x}_{i,j,1}^T, \ldots, \mathbf{x}_{i,j,n_t}^T \right]^T
\end{equation}

\begin{equation}
    Y = \left\{ \mathbf{y}_{1,1}, \mathbf{y}_{1,2}, \ldots, \mathbf{y}_{n_\mathrm{r}, n_\mathrm{s}} \right\}
\end{equation}
With this flattened input, a k-means clustering algorithm as described in~\cite{macqueen1967some} can be applied to obtain the cluster means, which are then used as the virtual targets.
Figure~\ref{fig:kmeans_clustering} illustrates the result of the k-means clustering for two dimensions.
This only serves as a visualization of the clustering process and does not represent the actual dimensions of the input data, since the input data, i.e., the flattened trajectories, are of size $n_t  d$ instead of two dimensions.

By choosing a desired $n_\mathrm{v}$, the number of virtual targets, the algorithm assigns each trajectory of the $n_\mathrm{r}  n_\mathrm{s}$ samples to one of the $n_\mathrm{v}$ clusters $C_i$ such that the sum of the squared Euclidean distances between the samples and the cluster means is minimized:
\begin{equation}
    \begin{aligned}
    \mathbf{c}_1^*, \ldots, \mathbf{c}_{n_\mathrm{v}}^* = \argminE_{\mathbf{c}_1, \ldots, \mathbf{c}_{n_\mathrm{v}}}& \sum_{i=1}^{n_\mathrm{v}} \sum_{\mathbf{y} \in C_i} ||\mathbf{y} - \mathbf{c}_i||^2 \label{eq:kmeans_objective} \\
    \mathrm{s.t.}\quad&\forall i \neq j:\quad C_i \cap C_j = \emptyset\\
    &~\bigcup_{i=1}^{n_\mathrm{v}} \quad C_i = Y
    \end{aligned}
\end{equation}
The result of the clustering process is a set of $n_\mathrm{v}$ optimal cluster means $\mathbf{c}_1^*, \dots, \mathbf{c}_{n_\mathrm{v}}^*$, each consisting of dimensions $(n_t  d, 1)$, which can then be reshaped to $n_\mathrm{v}$ trajectories with dimensions $(n_t, d)$ each:

\begin{equation}
    \mathbf{c}_i^* = \begin{bmatrix}
        \mathbf{c}_{i,1}^* & \ldots & \mathbf{c}_{i,d}^* \\
        \mathbf{c}_{i,d+1}^* & \ldots & \mathbf{c}_{i,2d}^* \\
        \vdots & \ddots & \vdots \\
        \mathbf{c}_{i,(n_t-1)  d + 1}^* & \ldots & \mathbf{c}_{i,n_t  d}^*
    \end{bmatrix}
    \label{eq:reshaping_clusters}
\end{equation}

\begin{equation}
    \phi_i(k  \Delta t) = \left( \mathbf{c}_{i,(k-1)  d + 1}^*, \ldots, \mathbf{c}_{i,k  d}^* \right)
    \label{eq:phi_i_t}
\end{equation}

\subsection{Renormalization}
As mentioned in Sec.~\ref{sec:application}, the samples are generated in a normalized coordinate system with axis limits of $[-1, 1]$.
To obtain the real-world positions and times of the virtual targets, the cluster means have to be renormalized to the original coordinate system, with the inverse of the normalization process mentioned in Sec.~\ref{sec:outlier_removal}.
After applying the renormalization, the trajectories of the virtual targets are obtained, which can be used for further analysis and downstream tasks.

\section{Data Generation}
\label{sec:data_generation}
Before training the CNFs model, data has to be obtained to train the model.
This data consists of trajectories of the targets.
If real data is available, it can be used to train the model instead of generating synthetic data.
However, since real trajectory data for missile targets is not available, synthetic data is generated for the two scenarios described in Sec.~\ref{sec:simple_target} and Sec.~\ref{sec:complex_target} using a Monte Carlo simulation and randomly generated target maneuvers.

\subsection{Simple Target}
\label{sec:simple_target}
As a practical illustration of the problem described in Sec.~\ref{sec:problem_statement}, consider a target flying in the horizontal plane, i.e., $\mathbf{x} \in \mathbb{R}^2$, with a constant velocity and performing random maneuvers. 
This scenario serves as an instructive example of a stochastically moving target.
Three different types of maneuvers are assumed: left turn, right turn, and straight flight. 
At the beginning of a trajectory (at the origin, flying northbound), the type and duration of the maneuver and the radius (bounded by the lateral acceleration) of the turn are randomly chosen from a uniform distribution with parameters depicted in Table~\ref{tab:maneuver_parameters} (reproduced from~\cite{CEAS-GNC-2024-103}).
After the maneuver is completed, new maneuvers are randomly chosen until the total duration of the trajectory is reached.
Since no additional parameters are required for this simple scenario, $\mathbf{\psi}$ is a vector of dimension zero ($\mathbf{\psi} \in \mathbb{R}^0$).
This simple maneuvering logic is maneuver-type independent, meaning that the target's behavior does not depend on its past behavior, which might not be realistic for all targets, but is sufficient for this simple scenario.
If this effect is to be considered, the model can be extended to include the past behavior of the target as additional conditioning variables as demonstrated by \cite{scholler2021flomo}.

\begin{figure}[h]
    \centering
    \includegraphics[width=0.8\textwidth, clip, trim=0 0 0 1.34cm]{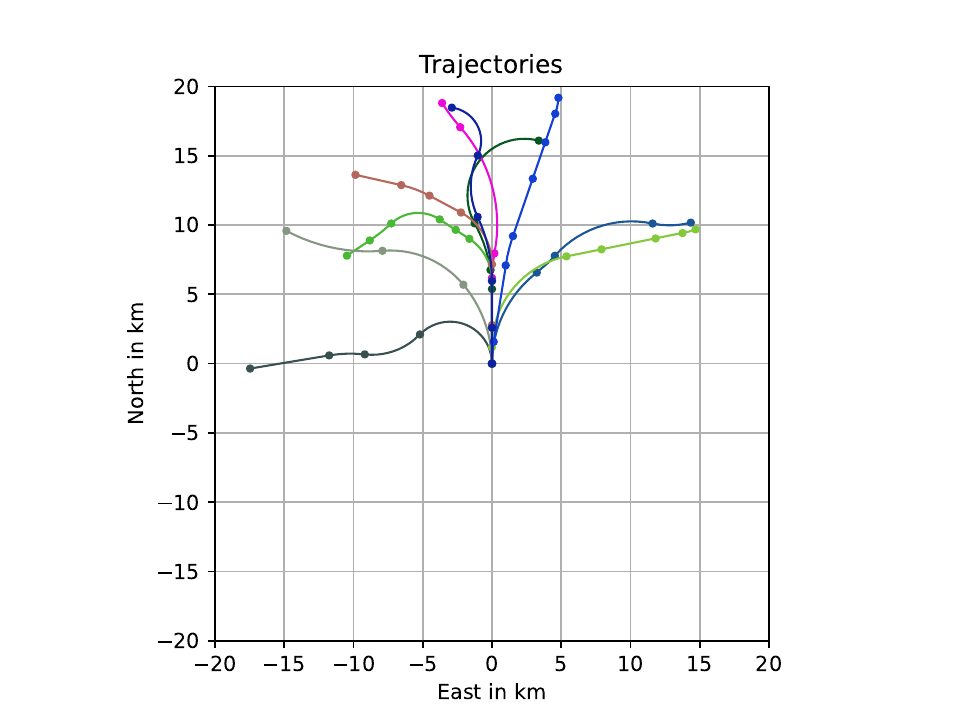}
    \caption{Ten randomly generated simple target trajectories with a duration of 100 s. Figure adapted from Schneider, M., Loureiro, R., Cunis, T., and Fichter, W., “Trajectory Prediction for Missile Targets: A Probabilistic Approach Using Machine Learning,” in CEAS EuroGNC Conference, 2024. Licensed under a Creative Commons Attribution 4.0 International License (CC-BY 4.0).}
    \label{fig:trajectories}
\end{figure}

Trajectories created with this approach are shown in Fig.~\ref{fig:trajectories} (adapted from~\cite{CEAS-GNC-2024-103}).
Each dot denotes the change of the maneuver.
Since the trajectories are simulated with a time discretization of 0.1~s, trajectory data can be obtained and saved for all the simulated time steps.

\begin{figure}
    \centering
    \begin{subfigure}[b]{0.32\textwidth}
        \includegraphics[width=\textwidth, clip, trim={3cm, 0cm, 2.4cm, 2cm}
        ]{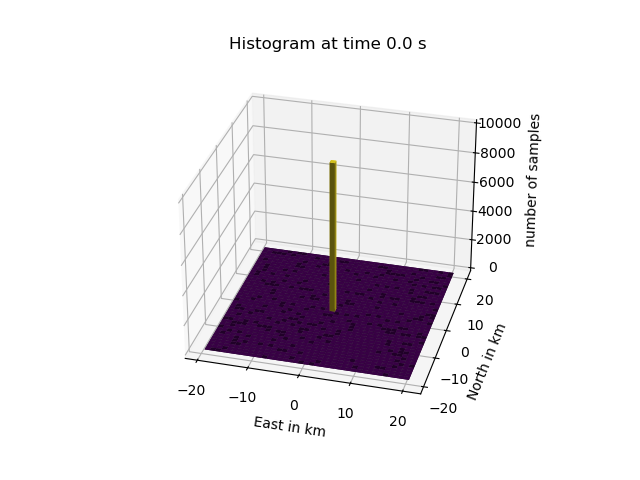}
        \caption{0~s}
        \label{fig:validation_0}
    \end{subfigure}
    \hfill
    \begin{subfigure}[b]{0.32\textwidth}
        \includegraphics[width=\textwidth, clip, trim={3cm, 0cm, 2.4cm, 2cm}
        ]{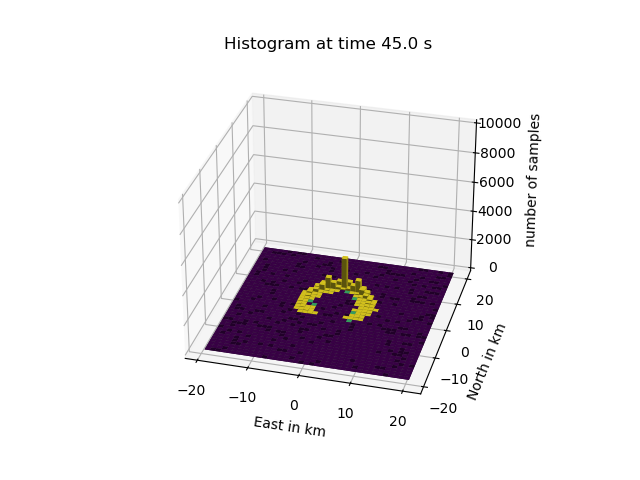}
        \caption{45~s}
        \label{fig:validation_45}
    \end{subfigure}
    \hfill
    \begin{subfigure}[b]{0.32\textwidth}
        \includegraphics[width=\textwidth, clip, trim={3cm, 0cm, 2.4cm, 2cm}
        ]{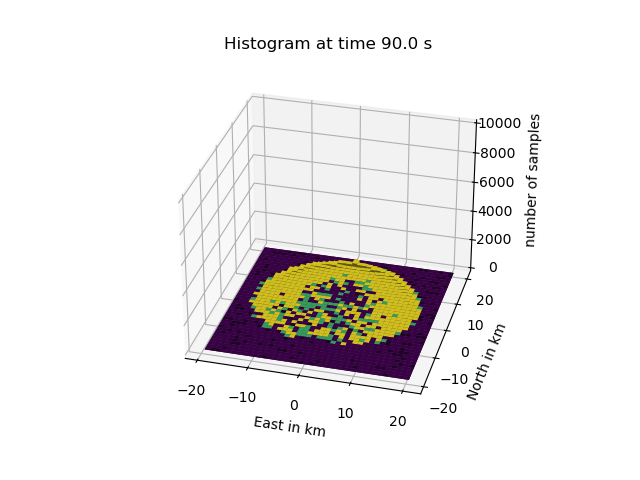}
        \caption{90~s}
        \label{fig:validation_90}
    \end{subfigure}
    \caption{Histograms of $10^4$ samples of the probability density function of the position of the target at different times obtained by Monte Carlo simulation. Figure reproduced from Schneider, M., Loureiro, R., Cunis, T., and Fichter, W., “Trajectory Prediction for Missile Targets: A Probabilistic Approach Using Machine Learning,” in CEAS EuroGNC Conference, 2024. Licensed under a Creative Commons Attribution 4.0 International License (CC-BY 4.0).}
    \label{fig:stochastic_validation}
\end{figure}

Figure~\ref{fig:stochastic_validation} (reproduced from~\cite{CEAS-GNC-2024-103}) depicts histograms of samples of the distribution of the target positions for different times.
The data was created with a computationally expensive Monte Carlo simulation.

\begin{table}[h]
    \centering
    \caption{Properties of the target maneuvers (reproduced from Schneider, M., Loureiro, R., Cunis, T., and Fichter, W., “Trajectory Prediction for Missile Targets: A Probabilistic Approach Using Machine Learning,” in CEAS EuroGNC Conference, 2024. Licensed under a Creative Commons Attribution 4.0 International License (CC-BY 4.0).).}
    \begin{tabular}{l l}
        \hline
        \hline
        \textbf{Property} & \textbf{Value} \\
        \hline
        Trajectory duration & $100~\text{s}$ \\
        Time discretization & $0.1~\text{s}$ \\
        Target speed & $200~\text{m/s}$ \\
        Minimum maneuver duration & $5~\text{s}$ \\
        Maximum maneuver duration & $50~\text{s}$ \\
        Minimum lateral acceleration & $3~\text{m/s}^2$ \\
        Maximum lateral acceleration & $20~\text{m/s}^2$ \\
        \hline\hline
    \end{tabular}
    \label{tab:maneuver_parameters}
\end{table}

\subsection{Complex Target}
\label{sec:complex_target}
In order to demonstrate the capabilities of the model, a more complex scenario is considered.
A ballistic target influenced by disturbances is considered, flying along a ballistic trajectory in three dimensions, i.e., $\mathbf{x} \in \mathbb{R}^3$.
Due to the disturbances, the target does not fly the ballistic trajectory exactly, but with some deviations, leading to a distribution of possible trajectories, which is to be modeled.
The ballistic trajectory is calculated according to Eq.~\eqref{eq:ballistic_model} in the North-East-Down frame.

\begin{equation}
    \begin{split}
        \dot{\mathbf{x}} &= \mathbf{v} \\
        \dot{\mathbf{v}} &= \mathbf{g} + \mathbf{d} + \mathbf{w} \\
    \end{split}
    \label{eq:ballistic_model}
\end{equation}
with 
\begin{align*}
    \mathbf{d} &= -\frac{1}{2  \xi}  \rho  ||\mathbf{v}||   \mathbf{v} \qquad  \text{with } \xi = \frac{m}{A  C_D} \\
    \mathbf{g} &= \left[0, 0, g\right]^\top \\
    \mathbf{w} & \sim \mathcal{N}(\mathbf{0}, \mathbf{\Sigma}) \\
\end{align*}

The change of the position $\mathbf{x}$ is equal to the velocity $\mathbf{v}$ and the change of the velocity $\mathbf{v}$ is equal to the sum of the gravitational acceleration $\mathbf{g}$, the deceleration $\mathbf{d}$ due to drag, and the disturbances $\mathbf{w}$.
The disturbance $\mathbf{w}$ is modeled as a zero-mean Gaussian noise with a covariance matrix $\mathbf{\Sigma}$.

The drag $\mathbf{d}$ is calculated with the ballistic coefficient $\xi$, which is the ratio of the mass $m$ of the target to the product of the cross-sectional area $A$ and the drag coefficient $C_D$ of the target.
The goal of the model is to calculate the distribution of the position $\mathbf{x}$ of the ballistic target after a certain time $t$.

Table~\ref{tab:maneuver_parameters_ballistic} (adapted from~\cite{CEAS-GNC-2024-103}) depicts the parameters of the ballistic trajectories used as training data.
The ballistic coefficient of the target is sampled from the uniform distribution
$\mathcal{U}(200, 800)~\mathrm{kg/m}^2$ to allow for the prediction of the distribution of the position of the target for any ballistic coefficient in the learned range.
This serves as an example of how additional parameters $\mathbf{\psi}$ can be included in the model to predict the distribution of the target's position for different parameters.
Thus, $\mathbf{\psi}$ is a vector of dimension one ($\mathbf{\psi} \in \mathbb{R}^1$) in this scenario.

\begin{table}[h]
    \centering
    \caption{Parameters of the ballistic trajectories used as training data (adapted from Schneider, M., Loureiro, R., Cunis, T., and Fichter, W., “Trajectory Prediction for Missile Targets: A Probabilistic Approach Using Machine Learning,” in CEAS EuroGNC Conference, 2024. Licensed under a Creative Commons Attribution 4.0 International License (CC-BY 4.0).).}
    \begin{tabular}{ll}
        \hline
        \hline
        \textbf{Parameter} & \textbf{Value} \\
        \hline
        Trajectory duration & 100 s \\
        Time discretization & 0.1 s \\
        Air density $\rho$ & 1.225 $\mathrm{kg}/\mathrm{m}^3$ \\
        Target ballistic coefficient & $\mathcal{U}(200, 800)$ $\mathrm{kg}/\mathrm{m}^2$ \\
        Target initial position & [0, 0, -1000] m \\
        Target initial velocity & [100, 0, 0] $\mathrm{m}/\mathrm{s}$ \\
        Disturbance covariance matrix $\mathbf{\Sigma}$ & $\mathbf{I}_3~\mathrm{m}^2/\mathrm{s}^4$ \\
        Gravitational acceleration $g$ & 9.81 $\mathrm{m}/\mathrm{s}^2$ \\
        \hline\hline
    \end{tabular}
    \label{tab:maneuver_parameters_ballistic}
\end{table}

Compared to the scenarios in Sec.~\ref{sec:simple_target}, the dynamics here differ dramatically:
\begin{enumerate}
    \item The target velocity is not constant but changes nonlinearly over time due to drag and the disturbance forces.
    \item The scenario takes place in three dimensions, instead of two dimensions.
    \item An additional parameter $\psi$ is added to the training data, namely the ballistic coefficient of the target.
\end{enumerate}

\begin{figure}[h]
    \centering
    \begin{subfigure}[b]{0.32\textwidth}
        \includegraphics[width=\textwidth, clip, trim={2cm, 0cm, 2cm, 2cm}]{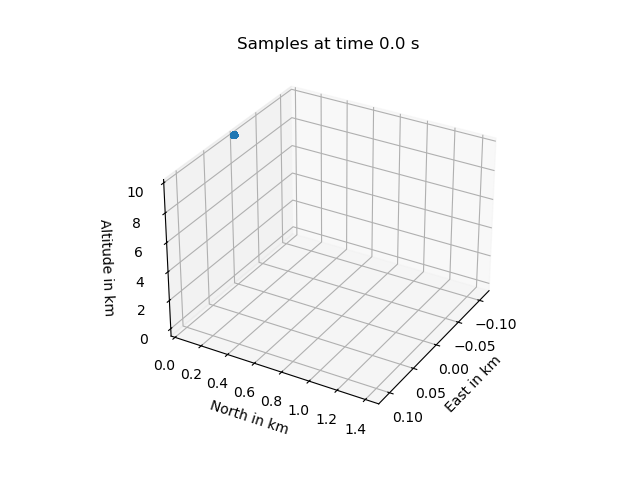}
        \caption{0~s}
        \label{fig:ballistic_trajectory_0}
    \end{subfigure}
    \hfill
    \begin{subfigure}[b]{0.32\textwidth}
        \includegraphics[width=\textwidth, clip, trim={2cm, 0cm, 2cm, 2cm}]{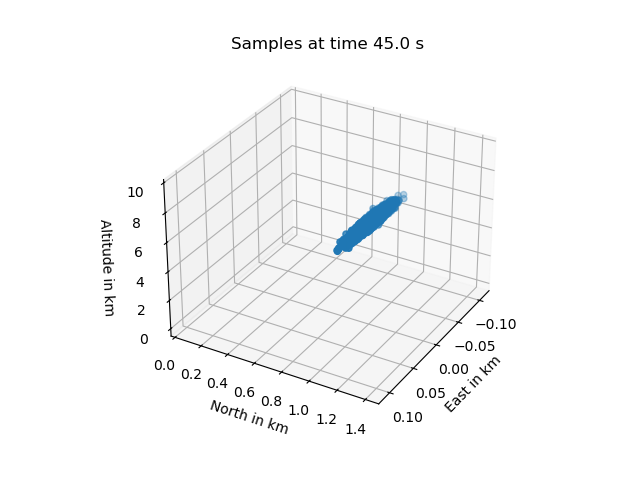}
        \caption{45~s}
        \label{fig:ballistic_trajectory_45}
    \end{subfigure}
    \hfill
    \begin{subfigure}[b]{0.32\textwidth}
        \includegraphics[width=\textwidth, clip, trim={2cm, 0cm, 2cm, 2cm}]{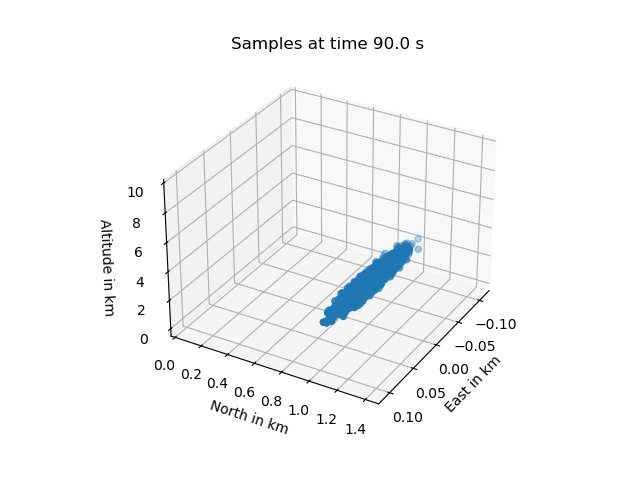}
        \caption{90~s}
        \label{fig:ballistic_trajectory_90}
    \end{subfigure}
    \caption{Monte Carlo simulation: samples of the probability density function of the position of the target with $\xi = 500~\frac{\mathrm{kg}}{\mathrm{m}^2}$ at different times. Figure reproduced from Schneider, M., Loureiro, R., Cunis, T., and Fichter, W., “Trajectory Prediction for Missile Targets: A Probabilistic Approach Using Machine Learning,” in CEAS EuroGNC Conference, 2024. Licensed under a Creative Commons Attribution 4.0 International License (CC-BY 4.0).}
    \label{fig:ballistic_trajectory}
\end{figure}

Figure~\ref{fig:ballistic_trajectory} (reproduced from~\cite{CEAS-GNC-2024-103}) depicts samples of the distribution of the target positions for different times and $\xi = 500~\mathrm{kg/m}^2$, obtained by Monte Carlo simulation.
It can be seen that the distribution of the target positions widens over time, due to the influence of the noise.

\section{Simulation Results}
\label{sec:results}
With the above-described CNFs approach, the distribution of the target's position can be modeled for any given time and target dynamics and then clustered to obtain a set of representative trajectories.
The results of the stochastic predictions by the CNFs are presented in Sec.~\ref{sec:stochastic_predictions}, followed by the generation of virtual target trajectories in Sec.~\ref{sec:generation_results}.

\subsection{Stochastic Predictions}
\label{sec:stochastic_predictions}

In the following, three different scenarios are considered: First, a stochastically moving target is examined in Sec.~\ref{sec:stochastic_maneuvers}.
Second, the application of the CNFs to predict trajectories of targets that are moving deterministically is presented in Sec.~\ref{sec:deterministic_maneuvers}.
Finally, the prediction of the position of a ballistic target with stochastic disturbances is examined in Sec.~\ref{sec:ballistic_targets}.
For all three scenarios, the same NNs and CNFs architecture are used, which are depicted in Tables~\ref{tab:nn_parameters} and~\ref{tab:cnf_parameters} (both adapted from~\cite{CEAS-GNC-2024-103}).

\subsubsection{Stochastic Maneuvers}
\label{sec:stochastic_maneuvers}
First, the scenario described in Sec.~\ref{sec:simple_target} is examined.
The latent dimension of the model, i.e., the dimension of the base distribution, is set to 2, which means that the model can learn the distribution of the position of the target in two dimensions, namely the $x$- and $y$-coordinate.
Thus, a two-dimensional Gaussian distribution is used as the base distribution that is transformed by the model to obtain the distribution of the position of the target.

\begin{figure}[h]
    \centering
    \begin{subfigure}[b]{0.32\textwidth}
        \includegraphics[width=\textwidth, clip, trim={3cm, 0cm, 2.4cm, 2cm}
        ]{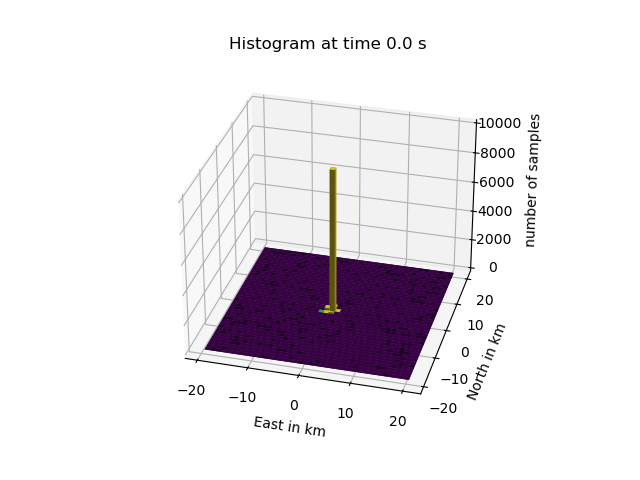}
        \caption{0~s}
        \label{fig:model_samples_0}
    \end{subfigure}
    \hfill
    \begin{subfigure}[b]{0.32\textwidth}
        \includegraphics[width=\textwidth, clip, trim={3cm, 0cm, 2.4cm, 2cm}
        ]{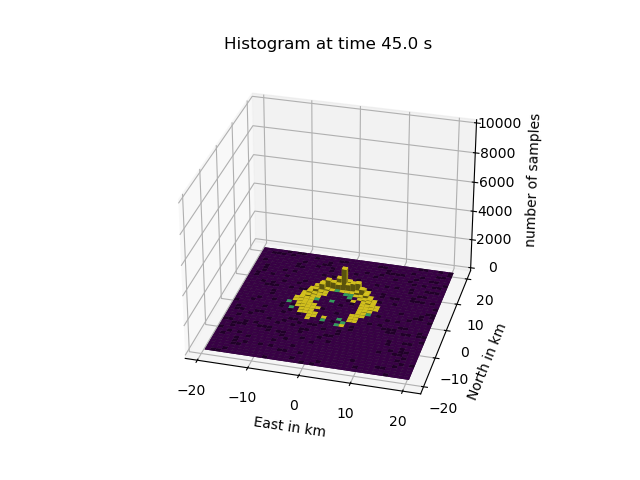}
        \caption{45~s}
        \label{fig:model_samples_45}
    \end{subfigure}
    \hfill
    \begin{subfigure}[b]{0.32\textwidth}
        \includegraphics[width=\textwidth, clip, trim={3cm, 0cm, 2.4cm, 2cm}
        ]{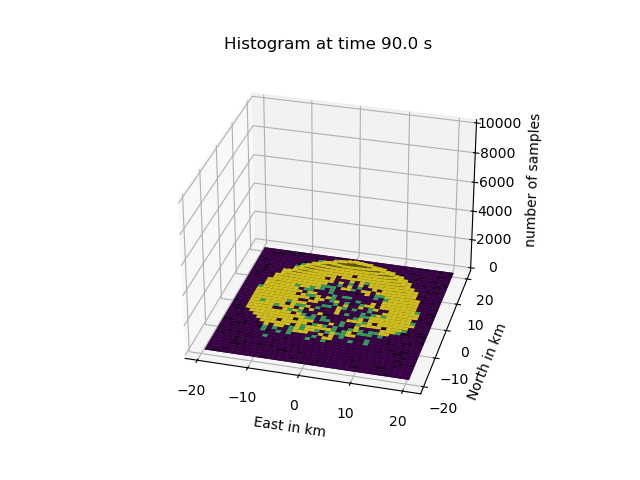}
        \caption{90~s}
        \label{fig:model_samples_90}
    \end{subfigure}
    \caption{Samples of the learned probability density function of the position of the target at different times. Figure reproduced from Schneider, M., Loureiro, R., Cunis, T., and Fichter, W., “Trajectory Prediction for Missile Targets: A Probabilistic Approach Using Machine Learning,” in CEAS EuroGNC Conference, 2024. Licensed under a Creative Commons Attribution 4.0 International License (CC-BY 4.0).}
    \label{fig:stochastic_samples}

\vspace{0.5cm}
    \centering
    \begin{subfigure}[b]{0.32\textwidth}
        \includegraphics[width=\textwidth, trim={0cm, 0cm, 0cm, 1.34cm}, clip]{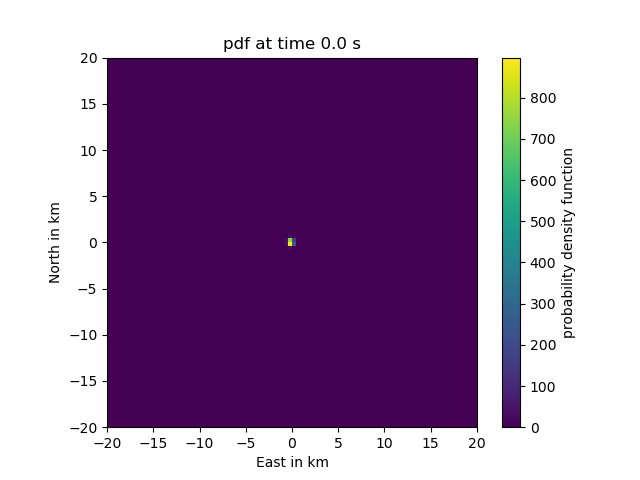}
        \caption{0~s}
        \label{fig:pdf_0}
    \end{subfigure}
    \hfill
    \begin{subfigure}[b]{0.32\textwidth}
        \includegraphics[width=\textwidth, trim={0cm, 0cm, 0cm, 1.34cm}, clip]{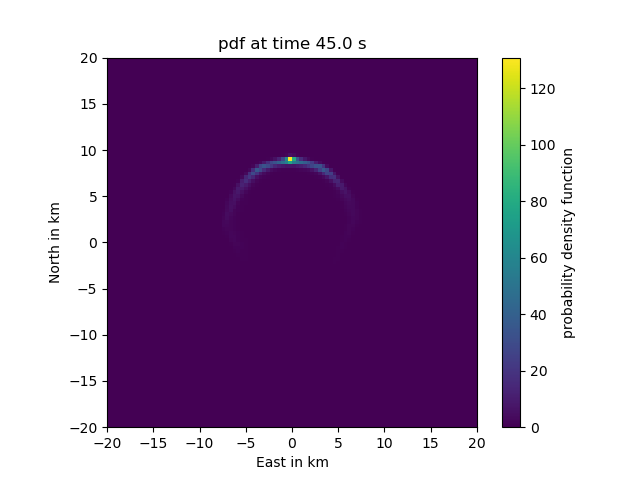}
        \caption{45~s}
        \label{fig:pdf_45}
    \end{subfigure}
    \hfill
    \begin{subfigure}[b]{0.32\textwidth}
        \includegraphics[width=\textwidth, trim={0cm, 0cm, 0cm, 1.34cm}, clip]{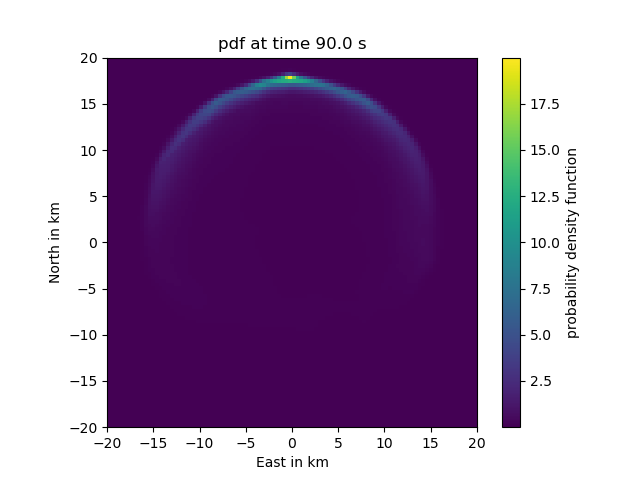}
        \caption{90~s}
        \label{fig:pdf_90}
    \end{subfigure}
    \caption{Learned probability density function of the position of the target at different times. Figure reproduced from Schneider, M., Loureiro, R., Cunis, T., and Fichter, W., “Trajectory Prediction for Missile Targets: A Probabilistic Approach Using Machine Learning,” in CEAS EuroGNC Conference, 2024. Licensed under a Creative Commons Attribution 4.0 International License (CC-BY 4.0).} 
    \label{fig:stochastic_pdf}
\end{figure}

After training the CNFs model for 1000 epochs (requiring 81 s of computation time\footnote{All computations were performed with a Ryzen 7 6800U processor with 16 GB RAM.}) on the generated data described in Sec.~\ref{sec:data_generation} and displayed in Fig.~\ref{fig:stochastic_validation}, the model is evaluated on the test data.
The results of the model evaluation are shown in Fig.~\ref{fig:stochastic_samples} (reproduced from~\cite{CEAS-GNC-2024-103}).
The figure depicts the absolute frequency from $10^4$ samples (as a representation of the PDF) of the position of the target at different times.
Figure~\ref{fig:stochastic_pdf} (reproduced from~\cite{CEAS-GNC-2024-103}) depicts the learned PDF of the position of the target at different times.

When comparing the results to the test data depicted in Fig.~\ref{fig:stochastic_validation}, it can be seen that the model is able to predict the position of the target quite well with a test loss of -2.85.
Furthermore, the calculation time for the model evaluation is independent of the time $t$ at which the distribution shall be predicted. 
Compared to the Monte Carlo simulation, the time to create $10^4$ samples is only about 0.17~s, compared to 6.64~s required for the Monte Carlo simulation for a simulated duration of 100~s.
Not only is the sampling much faster, the use of CNFs also allows for an exact computation of the PDF, which is not possible with the Monte Carlo trajectory generation method. 
Overall, we can conclude that CNFs are a suitable tool for modeling the distribution of the target's position.

\begin{table}
    \centering
    \caption{Parameters of the Neural Network (adapted from Schneider, M., Loureiro, R., Cunis, T., and Fichter, W., “Trajectory Prediction for Missile Targets: A Probabilistic Approach Using Machine Learning,” in CEAS EuroGNC Conference, 2024. Licensed under a Creative Commons Attribution 4.0 International License (CC-BY 4.0).).}
    \begin{tabular}{ll}
        \hline
        \hline
        \textbf{Parameter} & \textbf{Value} \\
        \hline
        Number of hidden layers & 2 \\
        Number of hidden units & 32 \\
        Activation function & ReLU \\
        Batch size & 1000 \\
        Learning rate & 0.003 \\
        Number of epochs & 1000 \\
        Optimizer & Adam \\
        Loss function & Negative log-likelihood \\
        \hline\hline
    \end{tabular}
    \label{tab:nn_parameters}
\end{table}

\begin{table}
    \centering
    \caption{Parameters of the Conditional Normalizing Flows model (adapted from Schneider, M., Loureiro, R., Cunis, T., and Fichter, W., “Trajectory Prediction for Missile Targets: A Probabilistic Approach Using Machine Learning,” in CEAS EuroGNC Conference, 2024. Licensed under a Creative Commons Attribution 4.0 International License (CC-BY 4.0).).}
    \begin{tabular}{ll}
        \hline
        \hline
        \textbf{Parameter} & \textbf{Value} \\
        \hline
        Number of flow layers $n_l$ & 4 \\
        Base distribution & Standard Gaussian \\
        Number of trajectories & $10^4$ \\
        Training data & 80\% \\
        Validation data & 20\% \\
        Noise injection standard deviation & 0.01 \\
        \hline\hline
    \end{tabular}
    \label{tab:cnf_parameters}
\end{table}

\subsubsection{Deterministic Maneuvers}
\label{sec:deterministic_maneuvers}
\begin{figure}
        \centering
        \begin{subfigure}[b]{0.32\textwidth}
            \includegraphics[width=\textwidth, clip, trim={3cm, 0cm, 2.4cm, 2cm}
            ]{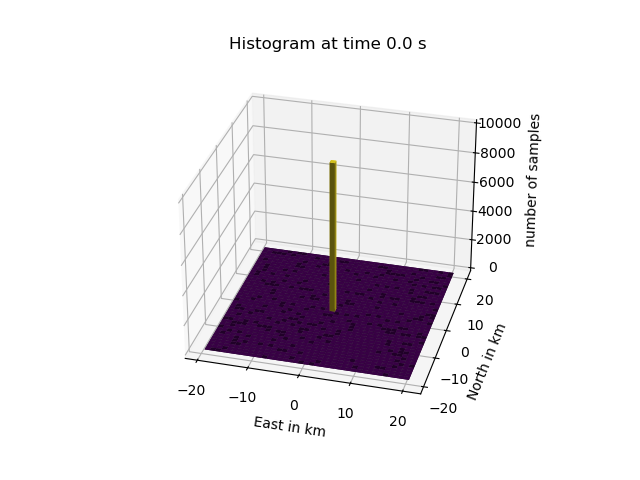}
            \caption{0~s}
            \label{fig:deterministic_validation_0}
        \end{subfigure}
        \hfill
        \begin{subfigure}[b]{0.32\textwidth}
            \includegraphics[width=\textwidth, clip, trim={3cm, 0cm, 2.4cm, 2cm}
            ]{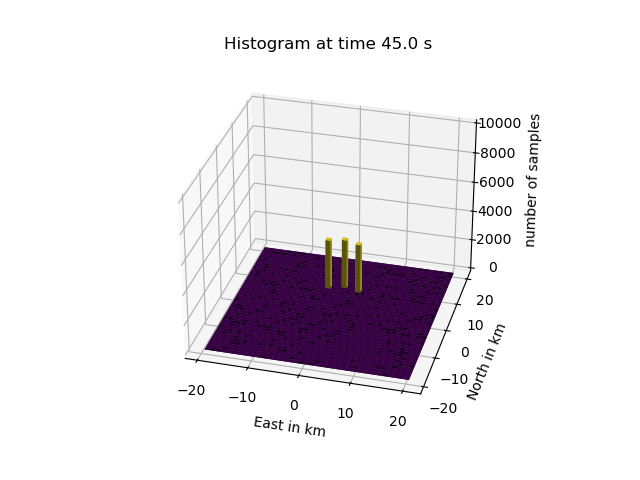}
            \caption{45~s}
            \label{fig:deterministic_validation_45}
        \end{subfigure}
        \hfill
        \begin{subfigure}[b]{0.32\textwidth}
            \includegraphics[width=\textwidth, clip, trim={3cm, 0cm, 2.4cm, 2cm}
            ]{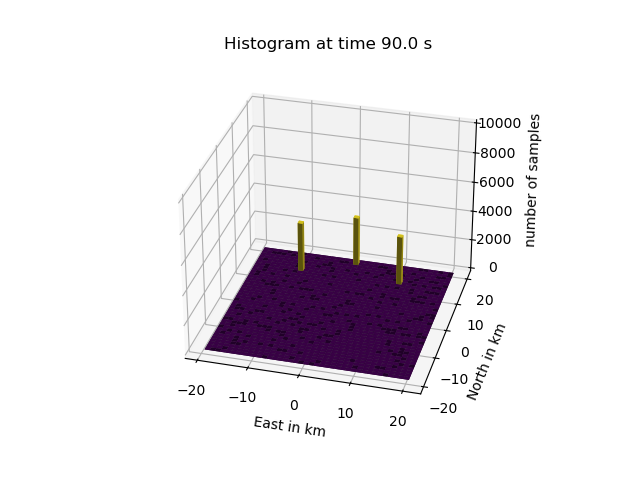}
            \caption{90~s}
            \label{fig:deterministic_validation_90}
        \end{subfigure}
        \caption{Test data: Histograms of samples of the probability density function of the position of the target at different times. Figure reproduced from Schneider, M., Loureiro, R., Cunis, T., and Fichter, W., “Trajectory Prediction for Missile Targets: A Probabilistic Approach Using Machine Learning,” in CEAS EuroGNC Conference, 2024. Licensed under a Creative Commons Attribution 4.0 International License (CC-BY 4.0).}
        \label{fig:deterministic_validation}

        \centering
        \begin{subfigure}[b]{0.32\textwidth}
            \includegraphics[width=\textwidth, clip, trim={3cm, 0cm, 2.4cm, 2cm}
            ]{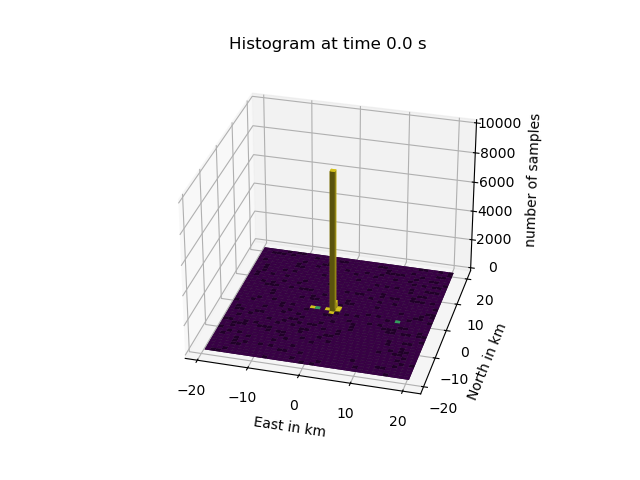}
            \caption{0~s}
            \label{fig:deterministic_model_samples_0}
        \end{subfigure}
        \hfill
        \begin{subfigure}[b]{0.32\textwidth}
            \includegraphics[width=\textwidth, clip, trim={3cm, 0cm, 2.4cm, 2cm}
            ]{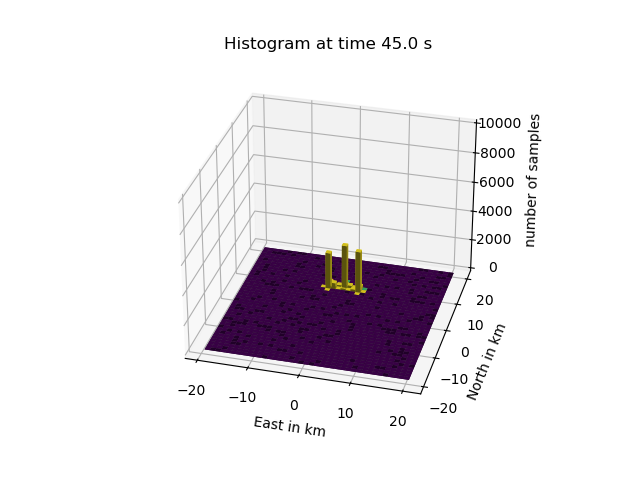}
            \caption{45~s}
            \label{fig:deterministic_model_samples_45}
        \end{subfigure}
        \hfill
        \begin{subfigure}[b]{0.32\textwidth}
            \includegraphics[width=\textwidth, clip, trim={3cm, 0cm, 2.4cm, 2cm}
            ]{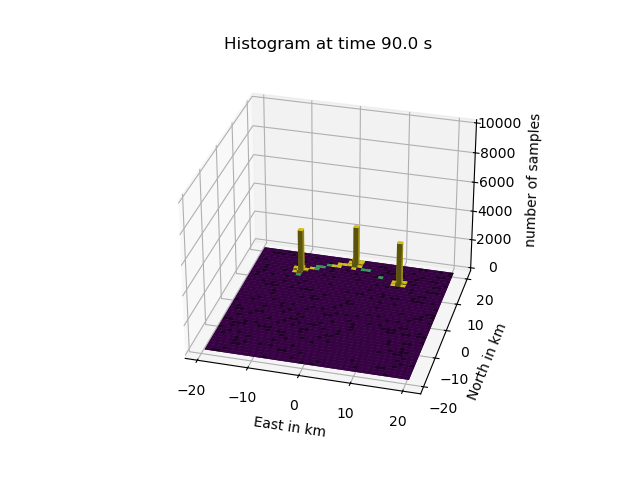}
            \caption{90~s}
            \label{fig:deterministic_model_samples_90}
        \end{subfigure}
        \caption{Learned distribution: Histograms of samples of the probability density function of the position of the target at different times. Figure reproduced from Schneider, M., Loureiro, R., Cunis, T., and Fichter, W., “Trajectory Prediction for Missile Targets: A Probabilistic Approach Using Machine Learning,” in CEAS EuroGNC Conference, 2024. Licensed under a Creative Commons Attribution 4.0 International License (CC-BY 4.0).}
        \label{fig:deterministic_samples}
    \end{figure}

To prove that a CNFs model with the same architecture can also be used for deterministic maneuvers, it is applied to predict the position of the target described in Sec.~\ref{sec:simple_target} after a deterministic maneuver of infinite duration.
The maneuvers tested were flying straight, a left turn, and a right turn, each with a fixed lateral acceleration of $3~\text{m/s}^2$.
In essence, this means the target can choose one of three different deterministic trajectories.
Compared to the previous scenario, the target dynamics have changed, thus new training data is created in a similar way as before, but with the new target dynamics.
Figure~\ref{fig:deterministic_validation} (reproduced from~\cite{CEAS-GNC-2024-103}) depicts the distribution of the target positions at different instances in the test data.

With the help of the new training data, the model is trained to predict the PDF of the position of the target at different times.
Figure~\ref{fig:deterministic_samples} (reproduced from~\cite{CEAS-GNC-2024-103}) depicts samples of the learned distribution of the predicted target positions for different times.
When comparing it to the test data in Fig.~\ref{fig:deterministic_validation}, it can be seen that the model is able to predict the position of the target quite well.
Nevertheless, there are some outliers, which are caused by the fact that the values of the learned PDF are not exactly zero.
Thus, the model predicts a very low, but non-zero, probability for the target to be at a position other than the three possible positions.

\subsubsection{Ballistic Targets}
\label{sec:ballistic_targets}
The same CNFs model as in the previous sections is used to predict the distribution of the ballistic trajectories described in Sec.~\ref{sec:complex_target}, but with a latent dimension (the dimension of the base distribution) of three, since the ballistic trajectories are simulated in three dimensions.
The training process was similar to the previous sections and required about 119~s.

With the help of the trained model, the distribution of the ballistic trajectory can be predicted without the need to simulate the ballistic trajectory multiple times. 
The required computation time for the prediction of $10^4$ samples using the CNFs is about 0.27~s, compared to the 178~s of the Monte Carlo simulation which was used to generate the training data.

\begin{figure}[h]
    \centering
    \begin{subfigure}[b]{0.32\textwidth}
        \includegraphics[width=\textwidth, clip, trim={2cm, 0cm, 2cm, 2cm}]{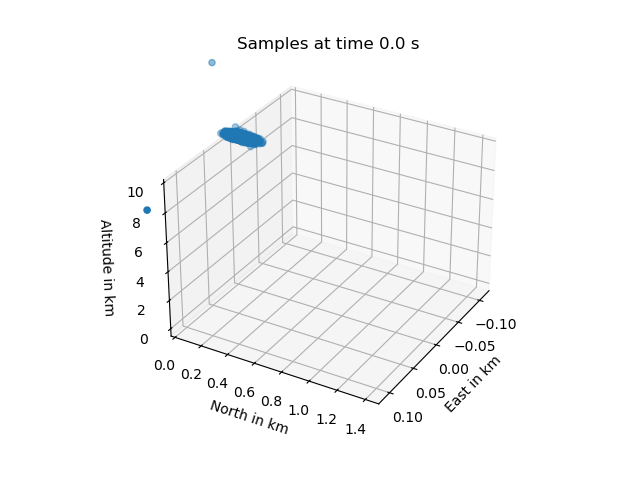}
        \caption{0~s}
        \label{fig:ballistic_trajectory_samples_0}
    \end{subfigure}
    \hfill
    \begin{subfigure}[b]{0.32\textwidth}
        \includegraphics[width=\textwidth, clip, trim={2cm, 0cm, 2cm, 2cm}]{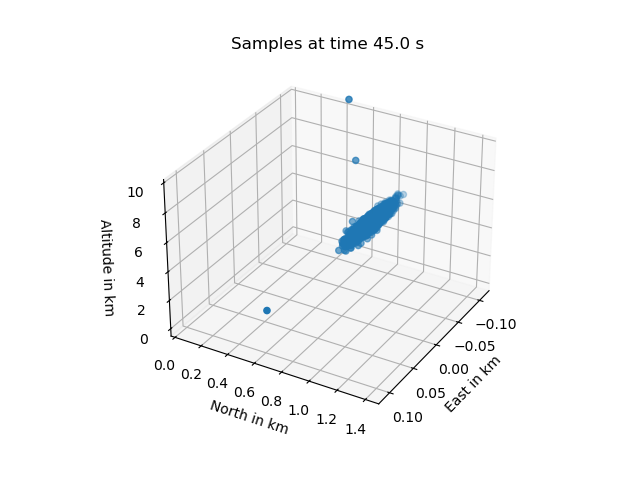}
        \caption{45~s}
        \label{fig:ballistic_trajectory_samples_45}
    \end{subfigure}
    \hfill
    \begin{subfigure}[b]{0.32\textwidth}
        \includegraphics[width=\textwidth, clip, trim={2cm, 0cm, 2cm, 2cm}]{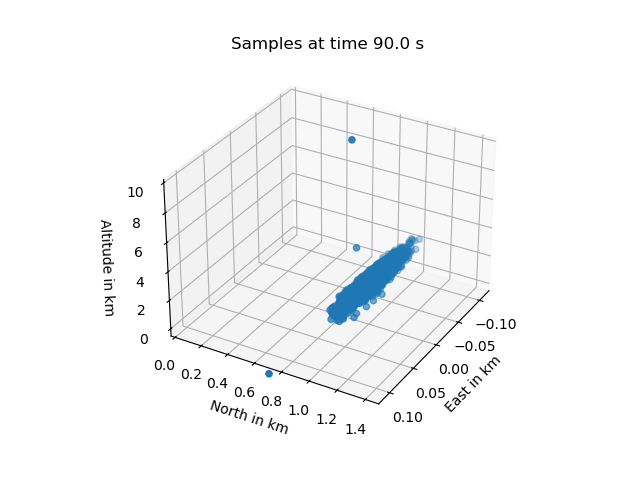}
        \caption{90~s}
        \label{fig:ballistic_trajectory_samples_90}
    \end{subfigure}
    \caption{Samples of the learned probability density function of the position of the target with the additional parameter $\psi = \xi = 500~\mathrm{kg/m}^2$ at different times. Figure reproduced from Schneider, M., Loureiro, R., Cunis, T., and Fichter, W., “Trajectory Prediction for Missile Targets: A Probabilistic Approach Using Machine Learning,” in CEAS EuroGNC Conference, 2024. Licensed under a Creative Commons Attribution 4.0 International License (CC-BY 4.0).}
    \label{fig:ballistic_trajectory_samples}
\end{figure}

Figure~\ref{fig:ballistic_trajectory_samples} (reproduced from~\cite{CEAS-GNC-2024-103}) depicts the samples of $p(\mathbf{x} \mid t, \psi)$ created for $t \in \{0, 45, 90\}$~s into the flight of the ballistic trajectory and $\psi = \xi = 500~\mathrm{kg/m}^2$.
When comparing the learned distribution to the test data depicted in Fig.~\ref{fig:ballistic_trajectory}, which was obtained through costly Monte Carlo simulation, we observe that the model is able to predict the distribution of the ballistic trajectory quite well. 
While there are some samples (e.g., three samples in Fig.~\ref{fig:ballistic_trajectory_samples_90}) that do not fit the training data, the overall shape of the distribution is similar to the test data depicted in Fig.~\ref{fig:ballistic_trajectory}, with a correct position of the mean of the distribution and a correct shape of the distribution.
Comparing this to the total of $10^4$ samples generated by the model, the number of such anomalies is considrabily low.

Only the distribution of the position of the target at 0~s is not predicted correctly: Whereas the mean of the distribution is predicted correctly, the shape of the distribution is not predicted correctly, being elongated along the direction of flight. 
A reason for this could be the training data: 
According to Table~\ref{tab:maneuver_parameters_ballistic}, all trajectories start at the position [0, 0, -1000] m, leading to a singularity in the PDF at this position, since the value of the PDF at this position is infinite to conserve an integral of 1.
The injection of noise mitigates this problem but does not solve it completely.

\subsection{Generation of Virtual Target Trajectories}
\label{sec:generation_results}
Through the use of the CNFs, stochastic predictions of the target's position can be made for any given time and target dynamics.
To make use of these predictions in applications such as guidance laws, path planning, etc., a set of representative trajectories is generated from the samples as described in Sec.~\ref{sec:generation}.
In the following, the results of the generation of virtual target trajectories are presented for various numbers of real targets and virtual targets, using the stochastically moving target from Sec.~\ref{sec:stochastic_maneuvers} as an example.
Table~\ref{tab:clustering_parameters} shows the parameters used in the clustering process.

\begin{table}[h]
    \centering
    \caption{Parameters of the clustering process.}
    \begin{tabular}{ll}
        \hline
        \hline
        \textbf{Parameter} & \textbf{Value} \\
        \hline
        Number of samples per target and time step & $200$ \\
        Number of time steps & 10 \\
        Tolerance to stop the clustering algorithm & $10^{-4}$ \\
        \hline\hline
    \end{tabular}
    \label{tab:clustering_parameters}
\end{table}

\subsubsection{Single Target}
\begin{figure}[h]
    \centering
    \begin{subfigure}[b]{0.49\textwidth}
        \includegraphics[width=\textwidth, clip, trim={1.2cm 0.4cm 1.2cm 0.9cm}]{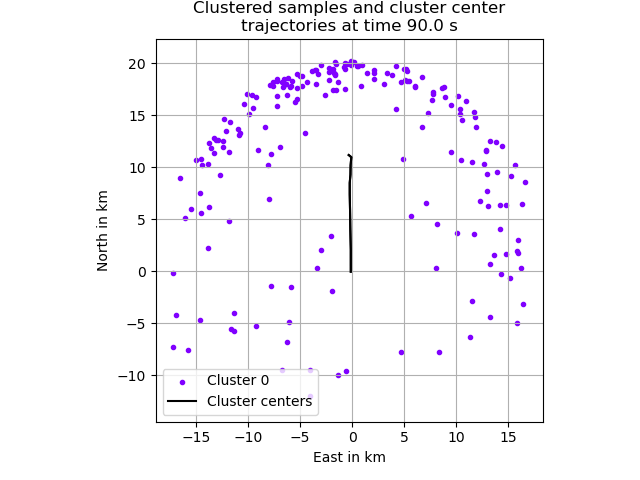}
        \caption{One virtual target}
        \label{fig:one_real_multiple_virtual_a}
    \end{subfigure}
    \hfill
    \begin{subfigure}[b]{0.49\textwidth}
        \includegraphics[width=\textwidth, clip, trim={1.2cm 0.4cm 1.2cm 0.9cm}]{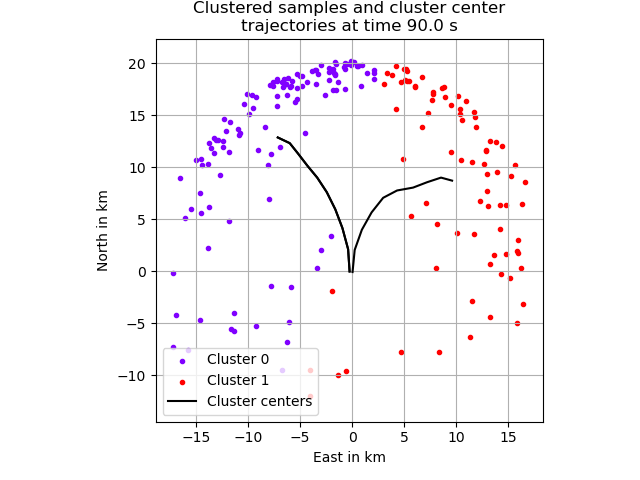}
        \caption{Two virtual targets}
        \label{fig:one_real_multiple_virtual_b}
    \end{subfigure}
    
    \vspace{0.5cm}
    
    \begin{subfigure}[b]{0.49\textwidth}
        \includegraphics[width=\textwidth, clip, trim={1.2cm 0.4cm 1.2cm 0.9cm}]{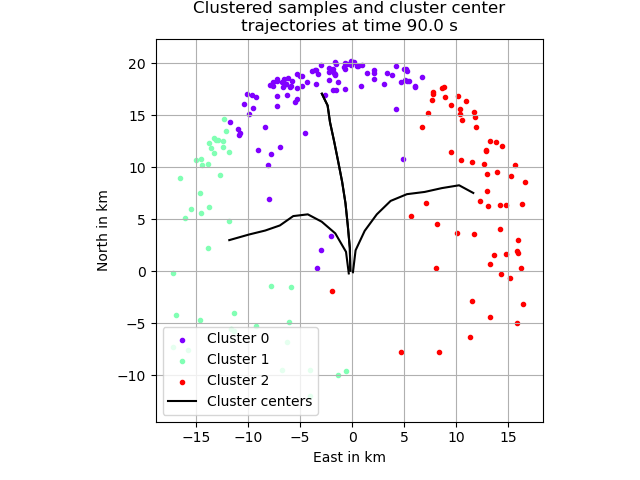}
        \caption{Three virtual targets}
        \label{fig:one_real_multiple_virtual_c}
    \end{subfigure}
    \hfill
    \begin{subfigure}[b]{0.49\textwidth}
        \includegraphics[width=\textwidth, clip, trim={1.2cm 0.4cm 1.2cm 0.9cm}]{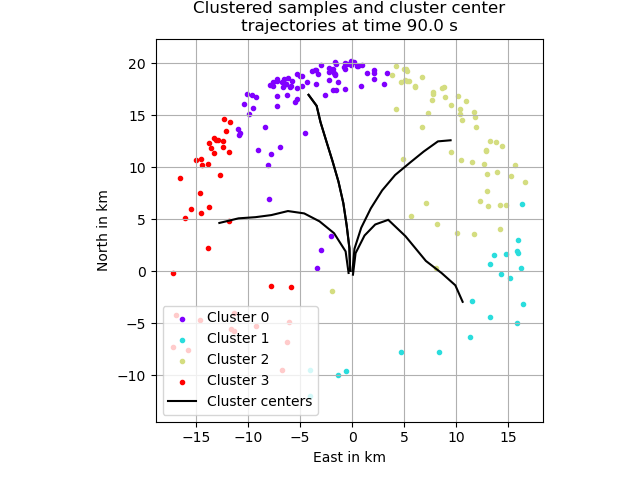}
        \caption{Four virtual targets}
        \label{fig:one_real_multiple_virtual_d}
    \end{subfigure}
    \caption{Prediction of virtual target trajectories for a single real target (initial position: $[0, 0]~\mathrm{km}$, flying northbound) and the respective clustered samples at 90~s.}
    \label{fig:one_real_multiple_virtual}
\end{figure}

The prediction for a single real target is shown in Fig.~\ref{fig:one_real_multiple_virtual} with one to four virtual targets.
When only one virtual target (Fig.~\ref{fig:one_real_multiple_virtual_a}) is predicted, the virtual target trajectory is similar to the average of the samples of the real target trajectories.
For applications where multiple virtual targets are required, the number of clusters can be increased.
The prediction of two virtual targets (Figures~\ref{fig:one_real_multiple_virtual_b}) splits the samples roughly symmetrically in the East-West direction, implying the possibility of a left and a right turn.
When three virtual targets are predicted (Fig.~\ref{fig:one_real_multiple_virtual_c}), the samples are assigned to three clusters: a left turn, a right turn, and a straight flight.
Interestingly, this concides with the multi-hypothesis approach in~\cite{schneider2022multi} where the target is assumed to perform one of multiple maneuvers, including a left turn, a right turn, and a straight flight.
However, the problem of defining the radii of the turns is avoided by the virtual target approach as the k-means clustering algorithm automatically determines the representative trajectories.
When four virtual targets are predicted (Fig.~\ref{fig:one_real_multiple_virtual_d}), the samples of the real target are clustered into four virtual targets, resulting in a more detailed prediction of the target's future position.

Overall, the trajectories cover a wide range of possible virtual target trajectories, without explicitly defining the maneuvers.
While the trajectories are not very smooth, the quality of the prediction can be improved by increasing the number of samples and the number of time steps.
Furthermore, smoothing techniques can be applied to the generated trajectories to obtain smoother trajectories if required.

\subsubsection{Multiple Targets}
As described in Sec.~\ref{sec:combination}, the approach can be extended to multiple real targets by generating samples for each target or type of target and then clustering the samples to obtain a set of representative trajectories.
In the following, the results for three real targets are shown, first for spatially separated targets and then for targets close to each other.
The total computation time to generate the 200 samples for each of the 10 time steps is 0.042~s. 
Clustering the samples takes about 0.1~s, regardless of the number of real and virtual targets.

\begin{figure}[h]
    \centering
    \begin{subfigure}[b]{0.49\textwidth}
        \includegraphics[width=\textwidth, clip, trim={1.2cm 0.4cm 1.2cm 0.9cm}]{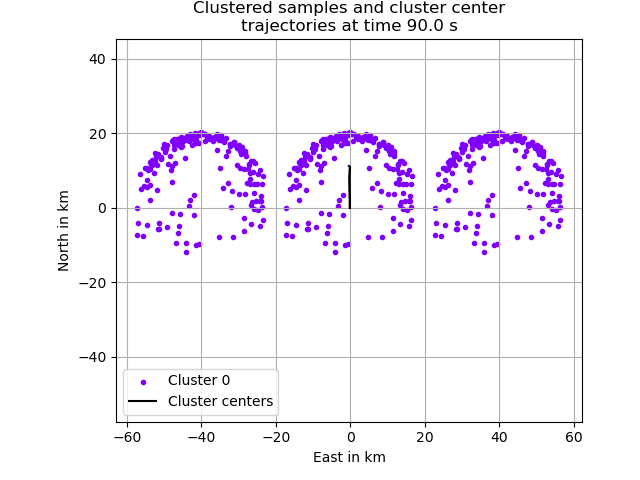}
        \caption{One virtual target}
        \label{fig:three_real_multiple_virtual_a}
    \end{subfigure}
    \hfill
    \begin{subfigure}[b]{0.49\textwidth}
        \includegraphics[width=\textwidth, clip, trim={1.2cm 0.4cm 1.2cm 0.9cm}]{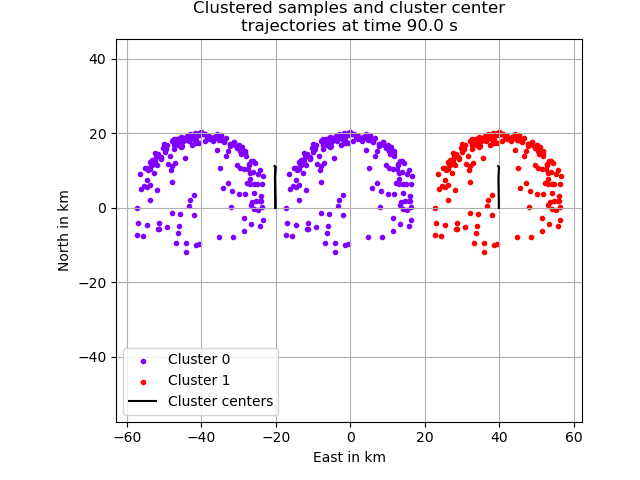}
        \caption{Two virtual targets}
        \label{fig:three_real_multiple_virtual_b}
    \end{subfigure}
    
    \vspace{0.5cm}
    
    \begin{subfigure}[b]{0.49\textwidth}
        \includegraphics[width=\textwidth, clip, trim={1.2cm 0.4cm 1.2cm 0.9cm}]{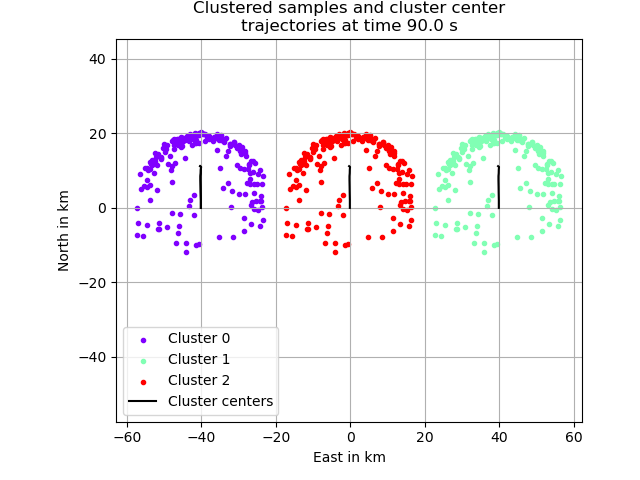}
        \caption{Three virtual targets}
        \label{fig:three_real_multiple_virtual_c}
    \end{subfigure}
    \hfill
    \begin{subfigure}[b]{0.49\textwidth}
        \includegraphics[width=\textwidth, clip, trim={1.2cm 0.4cm 1.2cm 0.9cm}]{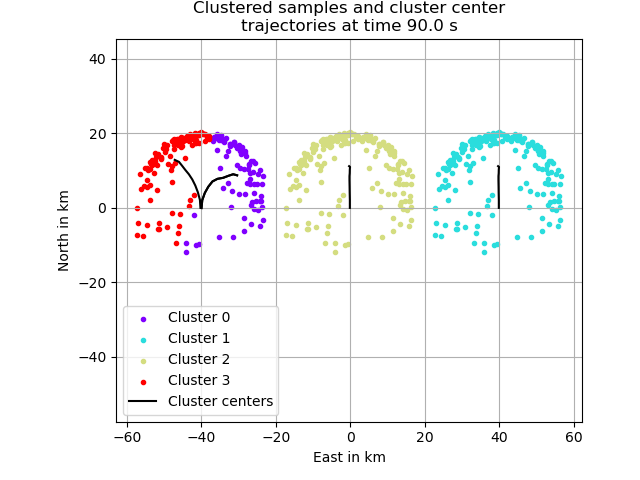}
        \caption{Four virtual targets}
        \label{fig:three_real_multiple_virtual_d}
    \end{subfigure}
    \caption{Prediction of virtual target trajectories for three real spatially separated targets (initial positions: $[0, -40]~\mathrm{km}, [0, 0]~\mathrm{km}, [0, 40]~\mathrm{km}$, flying northbound) and the respective clustered samples at 90~s.}
    \label{fig:three_real_multiple_virtual}
\end{figure}

\begin{figure}[h]
    \centering
    \begin{subfigure}[b]{0.49\textwidth}
        \includegraphics[width=\textwidth, clip, trim={1.2cm 0.4cm 1.2cm 0.9cm}]{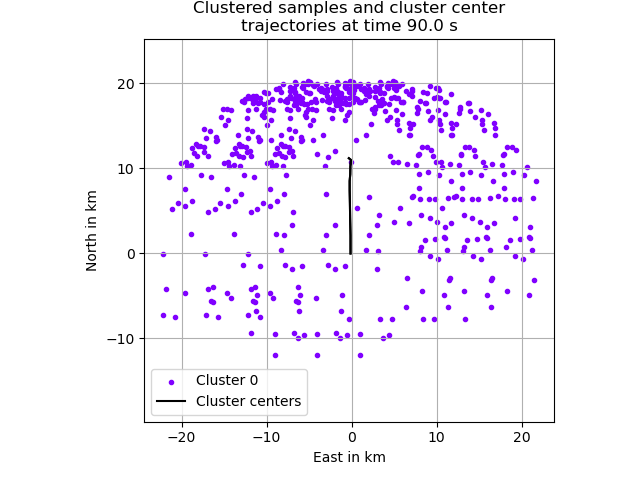}
        \caption{One virtual target}
        \label{fig:three_real_close_multiple_virtual_a}
    \end{subfigure}
    \hfill
    \begin{subfigure}[b]{0.49\textwidth}
        \includegraphics[width=\textwidth, clip, trim={1.2cm 0.4cm 1.2cm 0.9cm}]{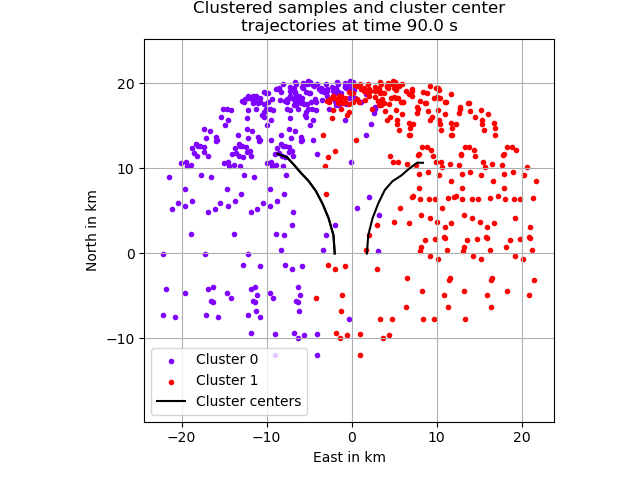}
        \caption{Two virtual targets}
        \label{fig:three_real_close_multiple_virtual_b}
    \end{subfigure}
    
    \vspace{0.5cm}
    
    \begin{subfigure}[b]{0.49\textwidth}
        \includegraphics[width=\textwidth, clip, trim={1.2cm 0.4cm 1.2cm 0.9cm}]{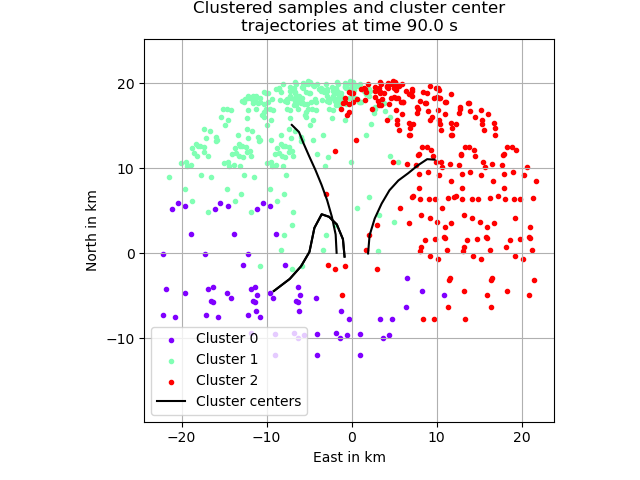}
        \caption{Three virtual targets}
        \label{fig:three_real_close_multiple_virtual_c}
    \end{subfigure}
    \hfill
    \begin{subfigure}[b]{0.49\textwidth}
        \includegraphics[width=\textwidth, clip, trim={1.2cm 0.4cm 1.2cm 0.9cm}]{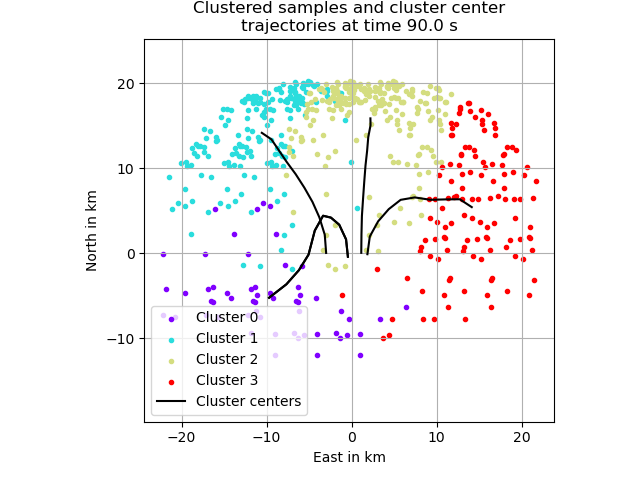}
        \caption{Four virtual targets}
        \label{fig:three_real_close_multiple_virtual_d}
    \end{subfigure}
    \caption{Prediction of virtual target trajectories for three real targets close to each other (initial positions: $[0, -5]~\mathrm{km}, [0, 0]~\mathrm{km}, [0, 5]~\mathrm{km}$, flying northbound) and the respective clustered samples at 90~s.}
    \label{fig:three_real_close_multiple_virtual}
\end{figure}

Figure~\ref{fig:three_real_multiple_virtual} shows the prediction for three real spatially separated targets with one to four virtual targets.
When only one virtual target is predicted for each real target, the virtual target trajectories are similar to the average of the real target trajectories (Fig.~\ref{fig:three_real_multiple_virtual_a}).
Increasing the number of virtual targets to two or three still results in the prediction of straight trajectories, just the position of the virtual targets is shifted (Figures~\ref{fig:three_real_multiple_virtual_b},~\ref{fig:three_real_multiple_virtual_c}).
When the number of virtual targets is higher than the number of real targets (Fig.~\ref{fig:three_real_multiple_virtual_d}), the samples of one real target are clustered into multiple virtual targets, similar to Fig.~\ref{fig:one_real_multiple_virtual}.
When the real targets are close to each other, as shown in Fig.~\ref{fig:three_real_close_multiple_virtual}, the distributions of the target positions overlap, leading to one left turn, one right turn, and one straight flight as virtual target trajectories.

\FloatBarrier

\section{Conclusion}
\label{sec:conclusion}
In this paper, an approach for the prediction of the representative trajectories of a stochastic target, denoted as virtual target trajectories, and the probability density function of its future position is presented.
The approach is based on the usage of Conditional Normalizing Flows, which are trained with the help of Monte Carlo simulation data, and time series k-means clustering to generate a set of representative trajectories.
The presented results demonstrate the method's effectiveness in predicting target positions for various scenarios, including targets with stochastic maneuvers, deterministic maneuvers, and ballistic trajectories with additional parameters.
The approach is target-agnostic and can be applied to different target types with appropriate training data.

Using Conditional Normalizing Flows, the position of the target at any given time can be predicted either by directly calculating the learned probability density function or by sampling from it. 
This approach is useful in various applications where predicting the target's position is essential.
Compared to Monte Carlo simulations, the approach can be computationally more efficient while also providing exact density evaluation: for the simple maneuver scenario creating $10^4$ samples the CNF required about 0.17~s compared to 6.64~s for the Monte Carlo reference, and for the ballistic scenario 0.27~s versus 178~s for Monte Carlo (see Sec.~\ref{sec:stochastic_predictions}).
These timing measurements illustrate the potential computational advantage of the learned model for large-sample evaluations.
Since most targets do not follow perfectly deterministic trajectories, the usage of a stochastic predictor can take the uncertainty of the target trajectory into account, potentially leading to more robust downstream applications.
When deterministic trajectories are required, the samples can be clustered to obtain a set of representative trajectories.
Thus, the presented approach allows to predict deterministic trajectories for stochastically moving targets, making it a versatile tool for trajectory prediction.
It can be used as a drop-in replacement for deterministic trajectory predictions used in areas such as guidance laws, path planning, and other applications where a prediction of the target's trajectory is needed.

\section*{Acknowledgments}
AI tools (namely ChatGPT and GitHub Copilot) were used to improve the readability of the manuscript and to assist with TikZ figures.

\bibliography{sample}

\end{document}